\documentclass[twocolumn]{article}
\usepackage{amsfonts}
\usepackage{amsmath}
\usepackage{cite}
\usepackage{graphicx} 
\usepackage{multirow}
\usepackage{booktabs}
\usepackage{subcaption}
\usepackage{hyperref}
\usepackage{authblk}
\usepackage{geometry} 
\usepackage{graphicx}   
\usepackage{subcaption}
\usepackage{multirow}
\usepackage{amsthm}
\usepackage{amsmath}
\newtheorem{definition}{Definition}
\newtheorem{theorem}{Theorem}
\newtheorem{lemma}{Lemma}   
\newtheorem{assumption}{Assumption}
\usepackage{makecell}

\usepackage[table]{xcolor}  
\usepackage{float}
\title{Orthogonal Soft Pruning for Efficient Class Unlearning}
\author{Qinghui Gong, Xue Yang\thanks{Corresponding author.}, Xiaohu Tang}
\affil{School of Information Science and Technology, Southwest Jiaotong University}
\date{}

\begin{document}

\maketitle
\begin{abstract}
Efficient and controllable data unlearning in federated learning remains challenging, due to the trade-off between forgetting and retention performance. Especially under non-independent and identically distributed (non-IID) settings, where deep feature entanglement exacerbates this dilemma. To address this challenge, we propose FedOrtho, a federated unlearning framework that combines orthogonalized deep convolutional kernels with an activation driven controllable one-shot soft pruning (OSP) mechanism. FedOrtho enforces kernel orthogonality and local–global alignment to decouple feature representations and mitigate client drift. This structural independence enables precise one-shot pruning of forgetting related kernels while preserving retained knowledge. FedOrtho achieves SOTA performance on CIFAR-10/100 and TinyImageNet with ResNet and VGG framework, verifying that FedOrtho supports class-, client-, and sample-level unlearning with over 98\% forgetting quality. It reduces computational and communication costs by 2-3 orders of magnitude in federated settings and achieves subsecond-level erasure in centralized scenarios while maintaining over 97\% retention accuracy and mitigating membership inference risks.
\end{abstract}

\section{Introduction}

\label{sec:intro}

Federated learning (FL)~\cite{konevcny2015federated,mcmahan2017communication,yang2019federated} enables collaboratively training a shared model without centralizing raw data, providing an effective pathway for privacy \cite{khalil2024dfml,nguyen2021federated,zhang2022federated}. With regulations such as the GDPR~\cite{union2023complete} and the CCPA~\cite{harding2019understanding}, the ``right to be forgotten” has become a critical demand of data subjects~\cite{garg2020formalizing,villaronga2018humans,mantelero2013eu}, compelling FL systems to support verifiable data unlearning~\cite{liu2025threats,xu2024machine}.


\begin{figure}[t]
  \centering
    \includegraphics[width=1.0\linewidth]{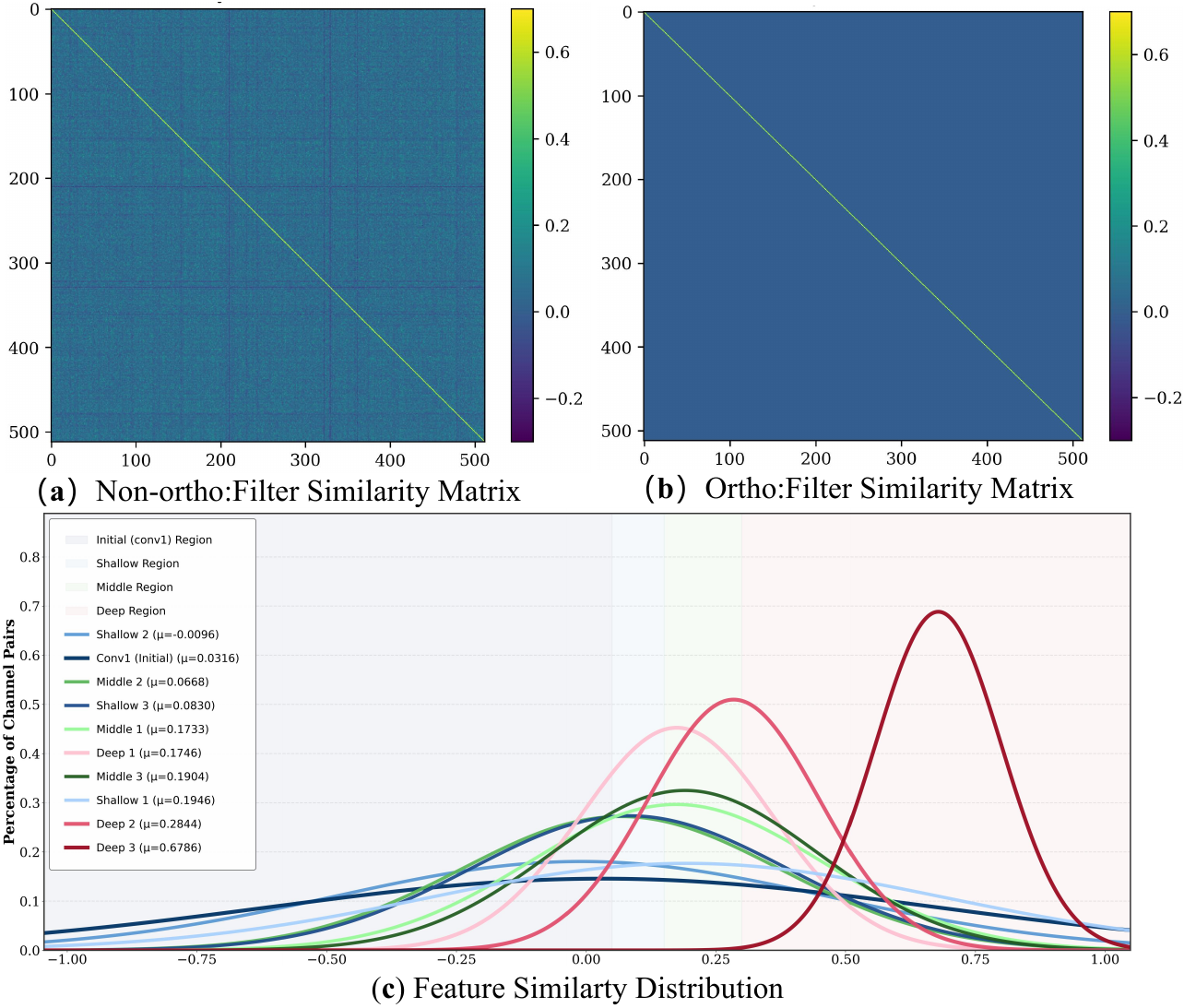}

   \caption{(a–b) Gram similarity heatmaps of \textit{layer4.2.conv2} show that orthogonal constraints significantly suppress the redundancy of conv kernels. (c) Feature coupling intensifies from shallow to deep layers.}
   \label{fig:ortho_vs_non_ortho}
\end{figure}

In this context, Federated Unlearning (FU)~\cite{halimi2022federated,liu2021federaser,wu2022federated,zhao2023federated,khalil2025not,zhong2025unlearning} is proposed to efficiently eliminate the impact of specific data from the global model without full retraining. Existing methods fall into two categories. Retraining-based methods~\cite{liu2021federaser,liu2020federated} could ensure complete unlearning, but they require repeating the entire federated training process, resulting in expensive communication and computational overhead, which makes them difficult to adapt to large-scale clients. Weight intervention-based methods~\cite{baumhauer2022machine,cao2023fedrecover,wang2023bfu}, which achieve fast unlearning through gradient reversal or clipping, but typically suffer from an accuracy imbalance between forgotten and retained sets.

We analyze and find that this imbalance stems from parameter redundancy and feature coupling within the model. As depth increases, deep convolutional (conv) kernels mix multi-sample features during representation, so forgetting target information induces non-targeted interference with retained knowledge \cite{wang2022federated}. As shown in Fig.~\ref{fig:ortho_vs_non_ortho}(c), the feature similarity of deep kernels is as high as 0.75, which intuitively confirms the significant coupling of feature representations. The non-IID data distribution unique to the federated setting further exacerbates this aliasing: local data bias causes deviations in the semantic directions of conv kernels across various clients, which destroys the semantic boundaries of features during global aggregation, intensifies deep feature coupling, and ultimately hinders the balance between complete forgetting and model stability.

Building on prior discussion, we note that the orthogonal constraint~\cite{wang2020orthogonal,choi2020role,achour2022existence} enforces independence among feature maps from different conv kernels, thereby enabling natural feature decoupling. As shown in Fig.~\ref{fig:ortho_vs_non_ortho}(a–b), before orthogonalization, the kernel similarity matrix contains large high-similarity regions, indicating strong semantic entanglement. After applying orthogonalization, these regions largely disappear, indicating individual kernels form distinct semantic response domains.

Inspired by this, we propose Efficient Federated Unlearning via Orthogonal Convolution and Adaptive Soft Pruning (FedOrtho), which achieves separation of semantics across kernels, yields a decoupled feature space, and thus effectively suppresses cross-sample and cross-class feature mixing. Moreover, leveraging the inherent sparsity of model parameters~\cite{jia2023model}, we perform one-shot, activation guided soft pruning to identify and attenuate kernels strongly tied to the data to be forgotten. This design prevents overlap of multi-source features within the same representation space. Furthermore, to precisely address feature drift in non-IID federated settings, we introduce local-global weight alignment during aggregation to stabilize the learned semantic space.

FedOrtho achieves SOTA performance on CIFAR10/100 and TinyImageNet using ResNet/VGG, with over 98\% forgetting quality. In federated settings, it is highly efficient, reducing both communication and computation costs by 2–3 orders of magnitude relative to SOTA methods. By combining orthogonal constraints, activation pruning, and model alignment, it cuts the retained-accuracy gap to under 3\% versus full retraining.

In summary, our contributions are listed as follows:
\begin{itemize}
\item We propose FedOrtho, an unlearning framework that decouples the feature space via conv kernel orthogonalization and integrates activation driven one-step soft pruning to locate unlearning kernels, erasing forgotten data while preserving retained knowledge.
\item We introduce orthogonal parameter alignment, which preserves the stability of the decoupled semantic space and thereby mitigates feature drift induced by non-IID data.
\item We conduct extensive experiments across three FU scenarios. Results validate that FedOrtho exhibits strong and sustainable non-IID robustness, along with excellent forgetting quality and retained model utility.
\end{itemize}

\section{Related Work}
\noindent \textbf{Federated Unlearning}\ \ \  FU extends traditional centralized unlearning~\cite{li2025causal,liu2025rethinking,felps2020class,graves2021amnesiac,tarun2023fast,bourtoule2021machine,chen2022graph} to FL settings, enabling client unlearning~\cite{gu2024unlearning}, class unlearning~\cite{li2023federated}, and sample unlearning~\cite{su2023asynchronous,wu2022federated} scenarios. The existing FU fall into two categories.
\textbf{(1) Retraining-based methods} ensure complete forgetting by removing the target data and retraining the model. For instance, \cite{liu2021federaser} retrains on the remaining clients to correct the global gradient, \cite{liu2022right} speeds up optimization via an improved Newton scheme, and \cite{su2023asynchronous} uses clustering to cut resource costs. Despite such efforts, both computation and communication overhead remain high in practice.
\textbf{(2) Weight-intervention methods}~\cite{ren2025advances,gao2024verifi,pan2025federated,zhang2023fedrecovery} exploit gradient exchange in FL to directly modify weights or the training pipeline for unlearning. FedEraser \cite{liu2021federaser} restores global model states to remove target-client influence. \cite{wu2022federated} accumulates historical updates and adds knowledge distillation. \cite{halimi2022federated} constrains gradient update ranges. 
And FCU \cite{deng2024enable} uses a contrastive loss to emulate a model absent the forgotten data. While these methods reduce training cost, they often rely on historical data, creating privacy risks and extra storage overhead.

\noindent \textbf{Unlearning via Weight Pruning}\ \ \ Weight-pruning-based forgetting methods in FU form a key class of weight manipulation techniques that prune parameters to suppress the influence of target data~\cite{baumhauer2022machine,guo2020dmcp,wang2022federated,jia2023model,fan2023salun}. \cite{wang2022federated} introduced a class discriminative pruning strategy for FU, but it incurs notable accuracy loss. \cite{golatkar2020forgetting} used the neural tangent kernel (NTK) to mitigate weight null-space issues and improve pruning specificity. \cite{foster2024fast} identified sensitive weights via Fisher information to refine pruning, but computing is costly. While pruning-based methods offer advantages in overhead, they share a key limitation: most target class unlearning scenarios and struggle to random or multi-round unlearning requests.

\section{Preliminary and Movation}
\subsection{Problem formulation of FU}

Let $W^{\mathrm{global}}$ denote the global weights collaboratively trained via federated optimization using FedAvg~\cite{mcmahan2017communication} over $K$ clients $\{C_k\}_{k=1}^{K}$:

\begin{equation}
W^{\mathrm{global}} = \arg\min_{W} \sum_{k=1}^{K} \frac{|D_k|}{|D|} \, \mathcal{L}(W; D_k),
\label{eq:fed_train}
\end{equation}
where $D_k$ is client $C_k$’s local dataset, $|D|=\sum_{k=1}^{K} |D_k|$, $\mathcal{L}$ is the client-side loss.
Given an unlearning request for a client or subset $D_u \subset D$, FU aims to produce a global model $W^{-u}$ whose performance on the retained data $D_r = D \setminus D_u$ matches that of $W^{r}$, the model retrained from on $D_r$. Re-training the model is defined as:

\begin{equation}
W^{r} = \arg\min_{W} \sum_{k: D_k \subset D_r} \frac{|D_k|}{|D_r|}\, \mathcal{L}(W; D_k).
\label{eq:fed_retrain}
\end{equation}

The ideal unlearning objective is formalized as:

\begin{equation}
P(y|x; W^{-u}) \approx P(y|x; W^{r}), 
\quad \forall x \in D_r,
\label{eq:retain_dist}
\end{equation}
that is, on the $D_r = D \setminus D_u$, the $W^{-u}$ output distribution matches the $W^{r}$, ensuring \textbf{Retention Consistency}.

\begin{equation}
P(y = y_{\text{true}}|x; W^{-u}) \approx 0, 
\quad \forall (x, y_{\text{true}}) \in D_u,
\label{eq:forget_dist}
\end{equation}
the model assigns near-zero probability to the true labels of the $D_u$, thereby ensuring a \textbf{Forgetting Guarantee}.

\subsection{Observation}

In FU experiments, we observe \textbf{a pronounced phenomenon of neuron feature overlap in deep conv layers:} a single kernel often responds to multiple classes, showing limited class selectivity and thus non-exclusive class representations.
To demonstrate this, we visualize activations of the 3rd and 8th kernels in a deep layer in Fig.~\ref{fig:activation} (left). Both yield positive activations for samples from the target and retained classes, revealing tightly coupled feature responses. This coupling means that changing the kernel weights will affect both the target and retained classes, lowering the unlearning precision.

\subsection{Motivation}

To address the problem of feature overlap in deep conv layers during FU, we introduce conv kernel orthogonality constraints to achieve feature decoupling, and provide theoretical proof from the perspective of dual space observation.

To clarify the mathematical nature of conv responses, we first define the conv functional in the dual space:
\begin{definition}[Conv Functional in Dual Space]
Let the input data space be $\mathbb{R}^D$ (where $D = C_{\text{in}} \cdot k^2$, $C_{\text{in}}$ is the number of input channels, and $k$ is the size of the conv kernel). The linear functionals in its dual space $(\mathbb{R}^D)^*$ satisfy: each conv kernel $\boldsymbol{w}_j \in \mathbb{R}^D$ uniquely determines a functional $f_j \in (\mathbb{R}^D)^*$. The response of $f_j$ to a flattened vector $\boldsymbol{x} \in \mathbb{R}^D$ of a $k \times k$ patch is $f_j(\boldsymbol{x}) = \boldsymbol{w}_j^\top \boldsymbol{x}$. The feature response intensity of the model to an image $X$ is quantified by the max value of functional responses across its $m$ patches, i.e., $\Phi_j(X) = \max_{1 \leq u,v \leq m} f_j(\boldsymbol{x}_{u,v})$.
\end{definition}

Based on the above definition of dual space functionals, we establish a direct connection from kernel orthogonality to irrelevant functional responses to feature decoupling through mathematical derivation.

\begin{theorem}[Feature Decoupling via Kernel Orthogonality]
Let the conv kernel matrix $\boldsymbol{W} \in \mathbb{R}^{C_{\text{out}} \times D}$ (where $C_{\text{out}}$ is the number of output channels) satisfy $\|\boldsymbol{W}\boldsymbol{W}^\top - \boldsymbol{I}\|_F \leq \varepsilon$ ($\boldsymbol{I}$ is the identity matrix, and $\varepsilon$ is the orthogonality deviation). Let $\mathcal{F}_{\text{target}}$ and $\mathcal{F}_{\text{retain}}$ be the sets of functionals for the target and retained classes, respectively.

For any $f_i \in \mathcal{F}_{\text{target}}$ and $f_j \in \mathcal{F}_{\text{retain}}$, the covariance of their responses to target class/sample images satisfies:
\begin{equation}
\begin{split}
\mathbb{E}[\Phi_i(X) \Phi_j(X)] 
&\leq B \cdot C \left( \varepsilon + (1+\varepsilon)\eta \right. \\
&\quad + \left. \sigma\sqrt{\frac{\log m_{\text{eff}}}{m_{\text{eff}}}} \right) =: \delta,
\end{split}
\end{equation}
where $B,C>0$ are constants, $\eta$ is the patch whitening error, $\sigma$ is the sub-Gaussian parameter of the functional, and $m_{\text{eff}}$ is the number of effective independent patches.

\textbf{Convergence}: As $\varepsilon \to 0$, $\eta \to 0$, and $m_{\text{eff}}$ becomes sufficiently large, $\delta \to 0$, indicating decoupling of features between the target and retained classes.
\end{theorem}

\begin{figure}[t]
  \centering
   \includegraphics[width=1.0\linewidth]{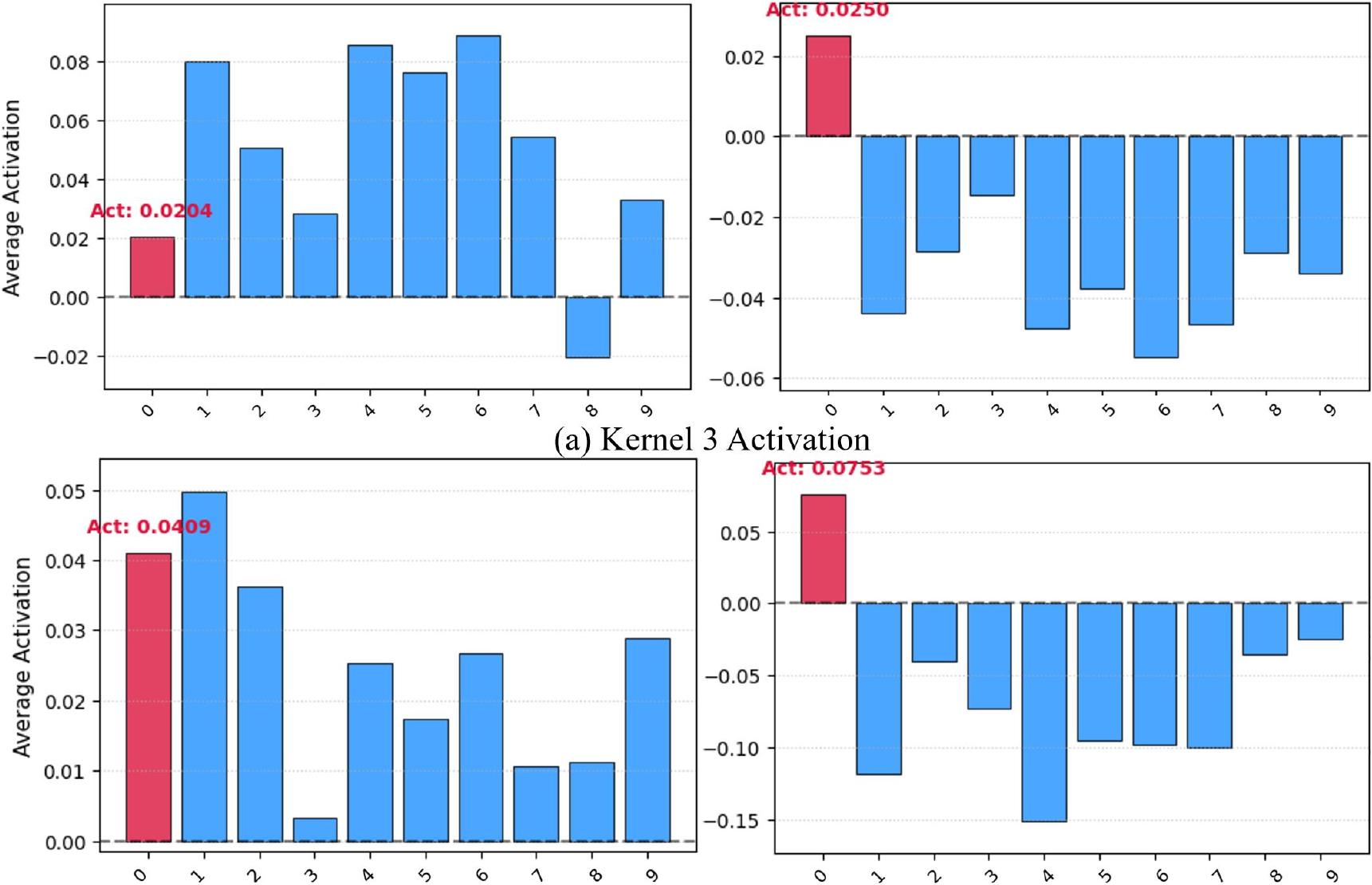}
   \caption{Activation differences across 10 classes in ResNet-50 \textit{layer4.2.conv2} (Class 0 as the unlearning target). (a)/(b) for 3rd/8th kernel. left/right for non-orthogonal/orthogonal model.}
   \label{fig:activation}
\end{figure}

This theorem mathematically proves that ``constraining conv kernel orthogonality" can achieve feature decoupling, providing a theoretical basis for subsequent method design. Experiments (the right panel of Fig.~\ref{fig:activation}) further verify that orthogonalized conv kernels exhibit strong activation for the target class and near-zero or negative activation for the retained classes, effectively resolving the feature overlap problem. Complete derivation of the theorem is provided in the appendix~\ref{app:decoupling_proof}.

\section{Method}

\begin{figure*}[t]
  \centering
   \includegraphics[width=1.0\linewidth]{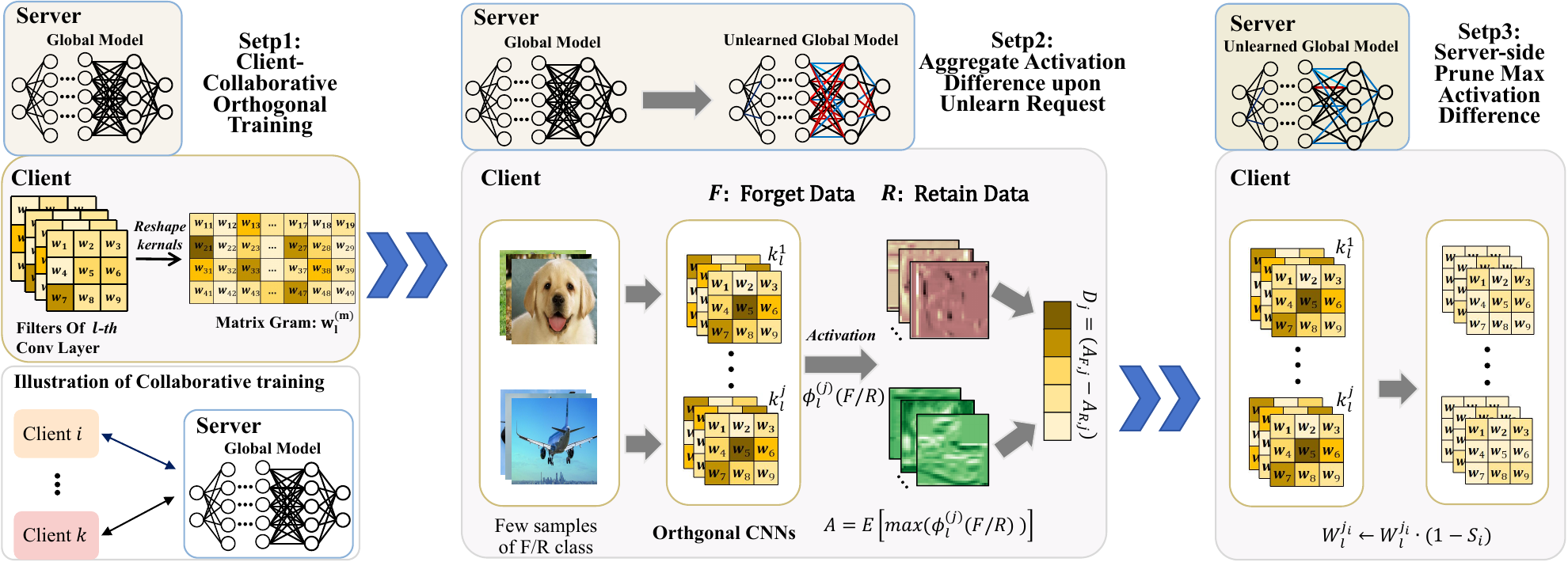}
   \caption{Illustration of the FedOrtho workflow. During collaborative training, clients train the model by introducing local orthogonal constraints and weight alignment. After an unlearning request is triggered, clients calculate local activation differences. The server aggregates differences from all clients and performs pruning to achieve knowledge unlearning without retraining the entire model.}
   \label{fig:overflow}
\end{figure*}

The proposed framework FedOrtho (Fig.~\ref{fig:overflow}) has two core modules: 
\textbf{(1) Federated Collaborative Orthogonal Training} implements a closed loop of ``local orthogonal optimization – global weight anchoring – secondary fusion optimization” to decorrelate conv kernels, align inter-client feature distributions, and induce sparsity to remove redundant kernels.
\textbf{(2) Distributed Activation-Driven Pruning} selects critical kernels from client-specific activation discrepancies, then the server aggregates global statistics to applies soft pruning without fine-tuning.

\subsection{Federated Collaborative Orthogonal Training}

\noindent \textbf{Client-Side Local Orthogonal Training}\ \ \ Each client $k$ trains on its local data $D_k$ with an orthogonality constraint to produce a local model $W_k$. To enforce inter kernel orthogonality, the weights of conv layer $l$, denoted $W^{(k)}_l$, is reshaped into a matrix $W^{(m,k)}_l \in \mathbb{R}^{C_{\text{out}} \times (C_{\text{in}} \cdot k_s^2)}$ to form the orthogonal sparse regularization term:

\begin{equation}
\mathcal{L}^{\text{local},k}_{\text{ortho-sparse}} = \left\| W^{(m,k)}_l \cdot W^{(m,k)\top}_l - I \right\|_F^2,
\end{equation}
where $I$ is the identity matrix and the Frobenius norm $\|\cdot\|_F$. This term augments the $\mathcal{L}_{\text{local},k}$:

\begin{equation}
\mathcal{L}_{\text{local},k} = \mathcal{L}_{\text{CE},k} + \lambda \cdot \mathcal{L}^{\text{local},k}_{\text{ortho-sparse}},
\end{equation}
where $\mathcal{L}_{\text{CE},k}$ is the cross-entropy on the local dataset $D_k$, and $\lambda$ tunes the strength of the $\mathcal{L}^{\text{local},k}_{\text{ortho-sparse}}$.

\vspace{4pt}
\noindent \textbf{Server-Side Global Reference Weights Generation}\ \ \ The server collects parameters from all clients, denoted by $\{W_k\}_{k\in\mathcal{K}}$, and computes weighted (based on data size) average to form the global model:

\begin{equation}
W^{\mathrm{global}}=\sum_{k\in\mathcal{K}} w_k\,W_k,\qquad
w_k=\frac{|D_k|}{\sum_{j\in\mathcal{K}}|D_j|},
\label{eq:global_aggregate}
\end{equation}
where $|D_{k}|$ is the sample count of $D_{k}$, $\mathcal{K}$ is the client set.

\vspace{4pt}
\noindent \textbf{Client-Side Second Stage Fusion Optimization}\ \ \ After receiving the global model $W^{\text{global}}$, each client adds $\mathcal{L}_{\text{align},k}$ to enforce consistency between local weights and the global reference, reducing non-IID drift and stabilizing aggregation. The corrected loss:

\begin{equation}
\mathcal{L}_{\text{total},k} = \mathcal{L}_{\text{CE},k} 
+ \lambda_1 \mathcal{L}^{\text{local},k}_{\text{ortho-sparse}} 
+ \lambda_2 \mathcal{L}_{\text{align},k},
\end{equation}
where $\mathcal{L}_{\text{align},k} = \|W_k - W^{\text{global}}\|_2^2$, $\lambda_1$ and $\lambda_2$ control the orthogonality sparsity constraint and the alignment strength, respectively. With the two stage optimization, the client weight $W_k$ achieves kernel independence and global consistency, forming the basis for accurate pruning.

\subsection{Distributed Activation Difference Statistics}
\noindent \textbf{Client-Side Local Activation Difference Computation}\ \ \ Each client $k$ estimates kernel-wise activation strength on its $D_k$. For the $j$-th kernel in layer $l$, it computes the spatial max activation on the local forget and retain sets:

\begin{align}
A^{\text{local},k}_{\text{target},j} &= \mathbb{E}_{x \in D_{k,u}} \left[ \max_{h,w} \phi^{(j)}_l(x)_{h,w} \right], \\
A^{\text{local},k}_{\text{retain},j} &= \mathbb{E}_{x \in D_{k,r}} \left[ \max_{h,w} \phi^{(j)}_l(x)_{h,w} \right],
\end{align}
where $D_{k,u}$ and $D_{k,r}$ are the local forget/retain sets. $\phi^{(j)}_l(x)_{h,w}$ is the activation at location $(h,w)$ in the feature map of kernel $j$ in layer $l$ for input $x$. The $\max_{h,w}$ captures the kernel’s peak response, and $\mathbb{E}$ averages over the dataset.
We measure the kernel’s association with the forgotten data by:

\begin{equation}
D^{\text{local},k}_j = A^{\text{local},k}_{\text{target},j} - A^{\text{local},k}_{\text{retain},j}.
\end{equation}

A larger $D^{\text{local},k}_j$ suggests kernel $j$ is forget dominant. Each client uploads $D^{\text{local},k}_j$ to the server as local evidence for global activation statistics.

\vspace{4pt}
\noindent \textbf{Server-Side Global kernel Aggregation}\ \ \ The server performs a weighted aggregation of each client’s local activation differences $D^{\text{local},k}_j$:

\begin{equation}
D^{\text{global}}_j = \sum_{k} w_k \cdot D^{\text{local},k}_j,
\end{equation}
where $D^{\text{global}}_j$ is the global activation difference for kernel $j$ tied to the data to be unlearned. The global differences over all kernels are sorted in descending order:

\begin{equation}
D^{(1)} \ge D^{(2)} \ge \cdots \ge D^{(C_{\text{out}})},
\end{equation}
given a pruning ratio $r \in (0,1]$, the pruned kernel $P$:

\begin{equation}
P = \{ j \mid D^{(j)} \ge D^{(\lceil r \cdot C_{\text{out}} \rceil)} \}.
\end{equation}

\noindent \textbf{Soft Pruning with Rank Based Adaptive Decay}\ \ \ To prevent accuracy drops from over pruning, we scale the pruning strength by kernel importance. Let the pruning set be
$P = {j_1, j_2, \dots, j_{N_p}}$, sorted in descending order by global activation discrepancy $D^{\text{global}}$. For each kernel $j_i$:

\begin{equation}
W_{l}^{j_i} \gets W_{l}^{j_i} \cdot (1 - S_i), \quad
S_i = \max\Big(\alpha, 1 - \frac{i}{N_p}\Big),
\end{equation}

where $i$ is the rank, $N_p$ is the number of pruned kernels, and $\alpha$ is the minimum pruning-strength threshold. This linear schedule aligns pruning with importance: higher ranked kernels receive stronger pruning, while lower ranked ones are lightly, yielding smoother degradation. The server then broadcasts $P$ and the corresponding strengths $S_i$ to all clients, which update their weights to perform unlearning.

\section{Experiment}

\begin{table*}[t]
    \caption{ Federated unlearning performance under different settings. For Client/Class/Sample Unlearning, we respectively forget the first client, first class, and 100 samples from one class; FT, NoT, FUSED rely on retraining for model restoration; FedOrtho$\dagger$ requires no retraining, with an excellent trade-off across all distributions. All table data are in percentage (\%). }
    \centering
    \resizebox{\linewidth}{!}{
    \begin{tabular}{c c ccc|ccc|ccc}
        \toprule
        \multicolumn{11}{c}{\textbf{Non-IID Scenario (Dirichlet=0.6)}} \\
        \midrule
        \multirow{2}{*}{\textbf{Model}} & \multirow{2}{*}{\textbf{Method}} 
        & \multicolumn{3}{c}{\textbf{Client Unlearning}} 
        & \multicolumn{3}{c}{\textbf{Class Unlearning}} 
        & \multicolumn{3}{c}{\textbf{Sample Unlearning}} \\ 
        \cmidrule(lr){3-5}  \cmidrule(lr){6-8}  \cmidrule(lr){9-11} 
        & & $A_{{D_u}} \downarrow$ & $A_{Te_{D_r}} \uparrow$& $\mathrm{MIA} \downarrow$ 
        & $A_{Te_{D_u}} \downarrow$ & $A_{Te_{D_r}} \uparrow$& $\mathrm{MIA} \downarrow$ 
        & $A_{{D_u}} \downarrow$ & $A_{Te_{D_r}} \uparrow$& $\mathrm{MIA} \downarrow$ \\
        \midrule

        \multirow{5}{*}{C100-R50} 
        & Retrain & 32.40±\footnotesize{0.89} & 48.35±\footnotesize{0.21}& 89.33±\footnotesize{0.56} & 0.00±\footnotesize{0.00} & 49.26±\footnotesize{0.42}& 0.67±\footnotesize{0.00} & 34.72±\footnotesize{1.03} & 49.29±\footnotesize{0.17}& 95.00±\footnotesize{0.45} \\ 
        & FT      & 40.20±\footnotesize{1.12} & \textbf{48.43}±\footnotesize{0.60} & 93.47±\footnotesize{0.68} & 5.33±\footnotesize{0.42} & \textbf{50.02}±\footnotesize{0.15} & 14.00±\footnotesize{0.91} & 41.33±\footnotesize{0.85} & \textbf{49.54}±\footnotesize{0.02} & 99.00±\footnotesize{0.33} \\
        & NoT     & 37.80±\footnotesize{2.95}& 45.13±\footnotesize{0.64} & 88.60±\footnotesize{1.93} & 7.67±\footnotesize{0.51} & 47.33±\footnotesize{0.89} & 4.33±\footnotesize{0.27} & 36.00±\footnotesize{1.14} & 47.96±\footnotesize{0.94} & 95.33±\footnotesize{0.52} \\
        & FUSED   & \underline{35.73}±\footnotesize{1.01} & 44.49±\footnotesize{0.83} & \underline{68.73}±\footnotesize{0.65} & \textbf{0.00}±\footnotesize{0.00} & 47.81±\footnotesize{1.05} & \textbf{3.00}±\footnotesize{0.21} & \underline{35.33}±\footnotesize{0.97} & \underline{48.36}±\footnotesize{1.11} & \textbf{73.33}±\footnotesize{0.69} \\
        & FedOrtho$\dagger$ & \cellcolor{gray!20}\textbf{32.60}±\footnotesize{0.02} & \cellcolor{gray!20}\underline{47.78}±\footnotesize{0.23} & \cellcolor{gray!20}\textbf{68.00}±\footnotesize{0.53} & \cellcolor{gray!20}\textbf{0.00}±\footnotesize{0.00} & \cellcolor{gray!20}\underline{48.37}±\footnotesize{0.04} & \cellcolor{gray!20}\underline{3.33}±\footnotesize{0.24} & \cellcolor{gray!20}\textbf{33.33}±\footnotesize{0.06} & \cellcolor{gray!20}48.17±\footnotesize{0.19} & \cellcolor{gray!20}\underline{76.00}±\footnotesize{0.61} \\
        \midrule

        \multirow{5}{*}{TIN-R50} 
        & Retrain & 26.26±\footnotesize{0.75} & 39.12±\footnotesize{0.09}& 78.27±\footnotesize{0.63} & 0.00±\footnotesize{0.00} & 39.07±\footnotesize{0.26}& 1.00±\footnotesize{0.00} & 30.00±\footnotesize{0.91} & 39.11±\footnotesize{0.04}& 83.00±\footnotesize{0.57} \\ 
        & FT      & 30.80±\footnotesize{1.02} & \textbf{39.92}±\footnotesize{0.65} & 83.53±\footnotesize{0.71} & 3.33±\footnotesize{0.38} & \textbf{39.53}±\footnotesize{0.03} & 12.33±\footnotesize{0.86} & 29.33±\footnotesize{0.88} & \textbf{39.45}±\footnotesize{0.79} & 93.67±\footnotesize{0.49} \\
        & NoT     & \underline{26.67}±\footnotesize{2.42}& 36.47±\footnotesize{0.06} & 77.33±\footnotesize{0.23} & 2.67±\footnotesize{0.32} & 36.59±\footnotesize{0.81} & 4.00±\footnotesize{0.67} & \underline{32.67}±\footnotesize{2.95} & 37.53±\footnotesize{0.87} & 82.00±\footnotesize{0.50} \\
        & FUSED   & 27.22±\footnotesize{0.83} & 34.72±\footnotesize{0.78} & \underline{61.93}±\footnotesize{0.56} & \textbf{0.00}±\footnotesize{0.00} & 37.58±\footnotesize{1.82} & \underline{3.33}±\footnotesize{0.23} & 33.33±\footnotesize{0.92} & \underline{38.79}±\footnotesize{0.31} & \underline{64.00}±\footnotesize{0.61} \\
        & FedOrtho$\dagger$ & \cellcolor{gray!20}\textbf{25.67}±\footnotesize{0.05} & \cellcolor{gray!20}\underline{38.26}±\footnotesize{0.14} & \cellcolor{gray!20}\textbf{61.67}±\footnotesize{0.53} & \cellcolor{gray!20}\textbf{0.00}±\footnotesize{0.00} & \cellcolor{gray!20}\underline{38.72}±\footnotesize{0.07} & \cellcolor{gray!20}\textbf{2.67}±\footnotesize{0.21} & \cellcolor{gray!20}\textbf{30.00}±\footnotesize{0.03} & \cellcolor{gray!20}38.66±\footnotesize{0.06} & \cellcolor{gray!20}\textbf{59.00}±\footnotesize{0.55} \\
        \midrule

        \multirow{5}{*}{C100-V16} 
        & Retrain & 29.73±\footnotesize{0.82} & 45.27±\footnotesize{0.14}& 84.07±\footnotesize{0.65} & 0.00±\footnotesize{0.00} & 45.51±\footnotesize{0.09}& 1.33±\footnotesize{0.00} & 34.67±\footnotesize{0.98} & 45.35±\footnotesize{0.11}& 89.00±\footnotesize{0.52} \\ 
        & FT      & 37.40±\footnotesize{1.05} & \textbf{46.05}±\footnotesize{0.45} & 89.40±\footnotesize{0.69} & 6.33±\footnotesize{0.45} & \textbf{45.85}±\footnotesize{0.88} & 19.33±\footnotesize{0.98} & 38.00±\footnotesize{1.01} & \textbf{45.95}±\footnotesize{0.05} & 97.33±\footnotesize{0.41} \\
        & NoT     & \textbf{30.60}±\footnotesize{4.14}& 42.63±\footnotesize{0.85} & 83.86±\footnotesize{2.06} & 3.00±\footnotesize{0.35} & 42.74±\footnotesize{0.79} & \textbf{2.00}±\footnotesize{0.16} & \underline{35.33}±\footnotesize{1.08} & 43.53±\footnotesize{0.83} & 88.33±\footnotesize{0.56} \\
        & FUSED   & 31.27±\footnotesize{0.84} & 43.11±\footnotesize{0.82} & \textbf{63.47}±\footnotesize{0.59} & \textbf{0.00}±\footnotesize{0.00} & 41.32±\footnotesize{0.84} & \underline{2.67}±\footnotesize{0.20} & 36.67±\footnotesize{0.43} & 40.26±\footnotesize{0.78} & \textbf{66.67}±\footnotesize{0.58} \\
        & FedOrtho$\dagger$ & \cellcolor{gray!20}\underline{30.73}±\footnotesize{0.05} & \cellcolor{gray!20}\underline{44.87}±\footnotesize{0.09} & \cellcolor{gray!20}\underline{65.40}±\footnotesize{0.62} & \cellcolor{gray!20}\textbf{0.00}±\footnotesize{0.00} & \cellcolor{gray!20}\underline{44.71}±\footnotesize{0.05} & \cellcolor{gray!20}3.00±\footnotesize{0.22} & \cellcolor{gray!20}\textbf{34.78}±\footnotesize{0.18} & \cellcolor{gray!20}\underline{44.88}±\footnotesize{0.11} & \cellcolor{gray!20}\underline{68.67}±\footnotesize{0.63} \\
        \midrule

        \multirow{5}{*}{TIN-V16} 
        & Retrain & 28.13±\footnotesize{0.79} & 37.14±\footnotesize{0.04} & 75.87±\footnotesize{0.61} & 0.00±\footnotesize{0.00} & 37.17±\footnotesize{0.14} & 0.67±\footnotesize{0.00} & 32.33±\footnotesize{0.94} & 37.32±\footnotesize{0.07} & 81.33±\footnotesize{0.53} \\ 
        & FT      & 32.53±\footnotesize{1.03} & \textbf{37.23}±\footnotesize{0.78} & 87.40±\footnotesize{0.72} & 3.67±\footnotesize{0.39} & \textbf{37.87}±\footnotesize{0.12} & 7.00±\footnotesize{0.64} & 35.00±\footnotesize{0.99} & \textbf{37.96}±\footnotesize{0.44} & 92.33±\footnotesize{0.47} \\
        & NoT     & 28.80±\footnotesize{1.19} & 35.27±\footnotesize{0.39} & 75.20±\footnotesize{0.84} & 3.00±\footnotesize{0.34} & 33.93±\footnotesize{0.76} & \underline{3.67}±\footnotesize{0.25} & \underline{29.60}±\footnotesize{0.85} & 34.87±\footnotesize{0.75} & 80.33±\footnotesize{0.34} \\
        & FUSED   & \underline{27.40}±\footnotesize{0.73} & \underline{36.95}±\footnotesize{0.80} & \textbf{57.00}±\footnotesize{0.52} & \underline{0.33}±\footnotesize{0.08} & 34.06±\footnotesize{0.79} & \textbf{3.33}±\footnotesize{0.20} & 33.51±\footnotesize{0.90} & \underline{36.87}±\footnotesize{0.73} & \textbf{65.00}±\footnotesize{0.56} \\
        & FedOrtho$\dagger$ & \cellcolor{gray!20}\textbf{26.67}±\footnotesize{0.08} & \cellcolor{gray!20}36.05±\footnotesize{0.07} & \cellcolor{gray!20}\underline{62.67}±\footnotesize{0.57} & \cellcolor{gray!20}\textbf{0.00}±\footnotesize{0.00} & \cellcolor{gray!20}\underline{36.73}±\footnotesize{0.08} & \cellcolor{gray!20}4.67±\footnotesize{0.31} & \cellcolor{gray!20}\textbf{28.33}±\footnotesize{0.14} & \cellcolor{gray!20}36.07±\footnotesize{0.06} & \cellcolor{gray!20}\underline{65.67}±\footnotesize{0.59} \\
        \midrule

        \multicolumn{11}{c}{\textbf{Non-IID Scenario (Dirichlet=0.3)}} \\
        \midrule
        \multirow{2}{*}{\textbf{Model}} & \multirow{2}{*}{\textbf{Method}} 
        & \multicolumn{3}{c}{\textbf{Client Unlearning}} 
        & \multicolumn{3}{c}{\textbf{Class Unlearning}} 
        & \multicolumn{3}{c}{\textbf{Sample Unlearning}} \\ 
        \cmidrule(lr){3-5}  \cmidrule(lr){6-8}  \cmidrule(lr){9-11} 
        & & $A_{{D_u}} \downarrow$ & $A_{Te_{D_r}} \uparrow$& $\mathrm{MIA} \downarrow$ 
        & $A_{Te_{D_u}} \downarrow$& $A_{Te_{D_r}} \uparrow$& $\mathrm{MIA} \downarrow$ 
        & $A_{{D_u}} \downarrow$& $A_{Te_{D_r}} \uparrow$& $\mathrm{MIA} \downarrow$ \\
        \midrule

        \multirow{5}{*}{C100-R50}
        & Retrain & 33.27±\footnotesize{0.91} & 47.15±\footnotesize{0.27} & 90.80±\footnotesize{0.64} & 0.00±\footnotesize{0.00} & 47.66±\footnotesize{0.89} & 0.67±\footnotesize{0.00} & 39.67±\footnotesize{1.02} & 48.09±\footnotesize{0.13} & 96.67±\footnotesize{0.43} \\ 
        & FT      & 41.47±\footnotesize{1.08} & \textbf{47.23}±\footnotesize{0.92} & 95.00±\footnotesize{0.58} & 5.67±\footnotesize{0.47} & \textbf{48.82}±\footnotesize{0.69} & 15.33±\footnotesize{0.92} & 42.67±\footnotesize{1.05} & \textbf{48.34}±\footnotesize{0.18} & 99.33±\footnotesize{0.32} \\
        & NoT     & \underline{35.93}±\footnotesize{2.86} & \underline{46.93}±\footnotesize{0.02} & 90.00±\footnotesize{0.66} & 8.33±\footnotesize{0.52} & 46.13±\footnotesize{0.85} & 5.67±\footnotesize{2.36} & 40.33±\footnotesize{1.03} & 46.76±\footnotesize{0.88} & 96.67±\footnotesize{0.45} \\
        & FUSED   & 37.00±\footnotesize{0.80} & 45.29±\footnotesize{0.87} & \textbf{70.21}±\footnotesize{0.63} & \textbf{0.00}±\footnotesize{0.00} & 46.61±\footnotesize{0.90} & \textbf{4.33}±\footnotesize{0.28} & \underline{36.67}±\footnotesize{0.97} & 47.16±\footnotesize{0.86} & \textbf{74.67}±\footnotesize{0.62} \\
        & FedOrtho$\dagger$ & \cellcolor{gray!20}\textbf{30.80}±\footnotesize{0.04} & \cellcolor{gray!20}46.58±\footnotesize{0.09} & \cellcolor{gray!20}\underline{79.47}±\footnotesize{0.71} & \cellcolor{gray!20}\textbf{0.00}±\footnotesize{0.00} & \cellcolor{gray!20}\underline{46.96}±\footnotesize{0.02} & \cellcolor{gray!20}\underline{4.67}±\footnotesize{0.30} & \cellcolor{gray!20}\textbf{35.00}±\footnotesize{0.02} & \cellcolor{gray!20}\underline{47.97}±\footnotesize{0.15} & \cellcolor{gray!20}\underline{77.33}±\footnotesize{0.65} \\
        \midrule

        \multicolumn{11}{c}{\textbf{IID Scenario}} \\
        \midrule
        \multirow{2}{*}{\textbf{Model}} & \multirow{2}{*}{\textbf{Method}} 
        & \multicolumn{3}{c}{\textbf{Client Unlearning}} 
        & \multicolumn{3}{c}{\textbf{Class Unlearning}} 
        & \multicolumn{3}{c}{\textbf{Sample Unlearning}} \\ 
        \cmidrule(lr){3-5}  \cmidrule(lr){6-8}  \cmidrule(lr){9-11} 
        & & $A_{{D_u}} \downarrow$ & $A_{Te_{D_r}} \uparrow$& $\mathrm{MIA} \downarrow$ 
        & $A_{Te_{D_u}} \downarrow$& $A_{Te_{D_r}} \uparrow$& $\mathrm{MIA} \downarrow$ 
        & $A_{{D_u}} \downarrow$ & $A_{Te_{D_r}} \uparrow$& $\mathrm{MIA} \downarrow$ \\
        \midrule

        \multirow{5}{*}{C100-R50}
        & Retrain & 31.47±\footnotesize{0.85} & 50.55±\footnotesize{0.07}& 87.87±\footnotesize{0.62} & 0.00±\footnotesize{0.00} & 50.46±\footnotesize{0.16}& 0.00±\footnotesize{0.00} & 35.00±\footnotesize{0.67} & 51.58±\footnotesize{0.08} & 93.67±\footnotesize{0.18}\\ 
        & FT      & 39.07±\footnotesize{1.06} & \textbf{51.63}±\footnotesize{1.03} & 92.05±\footnotesize{0.68} & 5.33±\footnotesize{0.43} & \textbf{51.12}±\footnotesize{0.58} & 12.67±\footnotesize{0.89} & 36.00±\footnotesize{0.97} & \textbf{51.74}±\footnotesize{0.06} & 99.33±\footnotesize{0.31} \\
        & NoT     & 36.67±\footnotesize{0.92} & \underline{50.64}±\footnotesize{0.18} & 87.33±\footnotesize{0.66} & 7.67±\footnotesize{0.50} & 50.53±\footnotesize{0.94} & 3.00±\footnotesize{2.80} & 35.33±\footnotesize{1.61} & \underline{50.86}±\footnotesize{0.97} & 94.00±\footnotesize{0.46} \\
        & FUSED   & \underline{34.60}±\footnotesize{0.40} & 45.69±\footnotesize{0.89} & \underline{67.37}±\footnotesize{0.57} & \textbf{0.00}±\footnotesize{0.00} & \underline{49.01}±\footnotesize{0.91} & \textbf{1.67}±\footnotesize{0.19} & \underline{35.00}±\footnotesize{1.00} & 49.56±\footnotesize{0.92} & \textbf{72.00}±\footnotesize{0.59} \\
        & FedOrtho$\dagger$ & \cellcolor{gray!20}\textbf{27.67}±\footnotesize{0.04} & \cellcolor{gray!20}49.98±\footnotesize{0.12} & \cellcolor{gray!20}\textbf{66.60}±\footnotesize{0.54} & \cellcolor{gray!20}\textbf{0.00}±\footnotesize{0.00} & \cellcolor{gray!20}49.77±\footnotesize{0.03} & \cellcolor{gray!20}\underline{2.00}±\footnotesize{0.22} & \cellcolor{gray!20}\textbf{32.67}±\footnotesize{0.03} & \cellcolor{gray!20}50.37±\footnotesize{0.06} & \cellcolor{gray!20}\underline{74.67}±\footnotesize{0.61} \\ 
        \bottomrule
    \end{tabular}
     }
    \label{tab:unlearning_iid_non_iid}
\end{table*}

\noindent \textbf{Datasets and Models}\ \ \ For experimental comparability and baseline consistency, we follow the dominant settings in the field. Datasets include CIFAR-10~\cite{krizhevsky2009learning}, CIFAR-100~\cite{krizhevsky2009learning}, and TinyImageNet~\cite{abai2019tinyimagenet}. Msodels adopt ResNet-18/50~\cite{cvpr/HeZRS16} and VGG-16~\cite{corr/SimonyanZ14a}.


\noindent \textbf{Federated Training Details}\ \ \ We adopt the FedAvg algorithm~\cite{mcmahan2017communication} for global model training across all datasets and model architectures. Specifically, we deploy \(N=100\) clients with non-IID partitions from a Dirichlet prior. The concentration \(\beta\) controls heterogeneity, using \(\beta=0.6\) (weak non-IID) and \(\beta=0.3\) (strong non-IID). Each round samples 10\% of clients for local updates over \(T=1000\) communication rounds. To improve conv kernel feature independence and stabilize aggregation under non-IID data, we add an orthogonality constraint (\(\lambda_{\text{ortho}}=0.1\)) and a weight-alignment constraint (\(\lambda_{\text{align}}=0.1\)), with weights selected via grid search (0.01-0.5). Specific hardware configurations and training parameters for both centralized and federated scenarios are provided in the Appendix~\ref{exp_setting}.

\noindent \textbf{Comparison}\ \ \ To ensure a comprehensive and scenario specific comparison, we evaluate FedOrtho against representative baselines in two settings, with all experiments run on three random seeds for result reliability. \textbf{Federated unlearning baselines}: Retrain only on the retain set $D_r$~\cite{liu2021federaser}, Fine-Tuning (FT) on $D_r$~\cite{cao2023fedrecover}. NoT~\cite{khalil2025not} negating weights of neurons, followed by FT on $D_r$. And FUSED~\cite{zhong2025unlearning} layer-sensitive adapters enabling reversible and efficient knowledge removal. \textbf{Centralized unlearning baselines}: SSD~\cite{foster2024fast}, Salun~\cite{fan2023salun}, EMNI~\cite{tarun2023fast}, and BTF~\cite{chundawat2023can}.

\noindent \textbf{Evaluation Metrics}\ \ \ 
\textbf{Retain Accuracy ($A_{Te_{D_r}}$)} reflects the model’s generalization performance on an independent test set that follows the same distribution as the retained data. \textbf{Forget Accuracy ($A_{D_u}$) }serves as an indicator of unlearning effectiveness, in sample or client unlearning scenarios, it is evaluated using the user revoked training subset $D_u$, while in class unlearning scenarios, it relies on a test set consistent with the distribution of the data requiring unlearning ($A_{Te_{D_u}}$).
Following~\cite{song2019privacy}, we use MIA~\cite{song2021systematic} for unlearning privacy evaluation. We define the Misclassification Rate to quantify privacy: \textbf{\(\text{MIA} = \frac{|D_{\text{TS}}|}{|D_u|} \times 100\%\) } where \(D_u\) is the unlearned dataset, \(D_{\text{TS}}\) denotes samples from \(D_u\) misclassified as training data under MIA.
In addition, we evaluated all federated methods using unlearning computation (FLOPs) and communication (Bytes) overhead, and further introduced unlearning time (s) for centralized settings.

\subsection{Experiment Result}

\begin{table*}[ht]
\caption{Comparative performance of the ResNet-50 model under the centralized scenario (forget the first class) on CIFAR-10, CIFAR-100, and Tiny ImageNet datasets.}
\centering
\setlength{\tabcolsep}{2.0pt} 
\fontsize{9pt}{10pt}\selectfont
\begin{tabular}{llcccc|cccc|cccc}
\toprule
\multirow{2}{*}{\textbf{Model}} & \multirow{2}{*}{\textbf{Method}} & \multicolumn{4}{c}{\textbf{CIFAR-10}} & \multicolumn{4}{c}{\textbf{CIFAR-100}} & \multicolumn{4}{c}{\textbf{Tiny ImageNet}} \\
\cmidrule(lr){3-6} \cmidrule(lr){7-10} \cmidrule(lr){11-14}
&  &  $A_{Te_{D_u}}{}_{\downarrow}$ & $A_{Te_{D_r}}{}_{\uparrow}$ & $\mathrm{Time}_{\downarrow}$ & $\mathrm{MIA}_{\downarrow}$ 
&  $A_{Te_{D_u}}{}_{\downarrow}$ & $A_{Te_{D_r}}{}_{\uparrow}$ & $\mathrm{Time}_{\downarrow}$ & $\mathrm{MIA}_{\downarrow}$ 
&  $A_{Te_{D_u}}{}_{\downarrow}$ & $A_{Te_{D_r}}{}_{\uparrow}$ & $\mathrm{Time}_{\downarrow}$ & $\mathrm{MIA}_{\downarrow}$ \\
\midrule
\multirow{7}{*}{Res50}
& Retrain       & 0.0\%  & 94.49\% & 10648.0s  & 0.00\%& 0.0\%  & 80.49\% & 11473.13s & 0.00\%& 0.0\%  & 65.32\% & 23105.83s & 0.00\%\\
& SSD           & 1.76\% & 90.23\% & 76.56s    & 5.80\%         & 3.67\%& \underline{78.15\%}& 74.42s    & 3.00\%         & 5.33\%& 61.87\% & 480.06s   & 4.00\%         \\
& EMNI          & 2.63\% & \underline{92.28\%}& 784.75s   & 3.00\%         & 4.0\%& 78.02\% & 795.32s   & 3.67\%         & 7.33\%& \underline{62.43\%}& 800.26s   & 3.33\%         \\
& Salun         & \textbf{0.0\%}& 87.49\% & \underline{43.07s}& \textbf{0.60\%} & \textbf{0.0\%}& 72.09\% & \underline{44.69s}& \textbf{0.00\%} & \textbf{0.0\%}& 59.62\% & \underline{87.04s}& \textbf{0.00\%} \\
& BTF           & 7.33\%& 86.34\% & 58.93s    & 3.67\%         & 6.00\%& 74.14\% & 59.41s    & 4.00\%         & 0.63\%& 54.25\% & 129.84s   & 4.67\%         \\
& FedOrtho$\dagger$ & \cellcolor{gray!20}\textbf{0.0\%}& \cellcolor{gray!20}\textbf{93.67}\% & \cellcolor{gray!20}\textbf{0.1584s} & \cellcolor{gray!20}\underline{2.00\%}& \cellcolor{gray!20}\textbf{0.0\%}& \cellcolor{gray!20}\textbf{79.70\%}& \cellcolor{gray!20}\textbf{0.1546s} & \cellcolor{gray!20}\underline{2.00\%}& \cellcolor{gray!20}\textbf{0.0\%}& \cellcolor{gray!20}\textbf{64.55\%}& \cellcolor{gray!20}\textbf{0.1676s} & \cellcolor{gray!20}\underline{2.67\%}\\
\bottomrule
\end{tabular}
\label{tab:unlearning_res50}
\end{table*}

\begin{table}[htbp]
  \caption{Comparison of continuous unlearning performance between non-IID and centralized settings (ResNet-50/CIFAR-100, five consecutive unlearned classes).}
  \centering
  \fontsize{9pt}{10pt}\selectfont
  \setlength{\tabcolsep}{5pt}
  \begin{tabular}{c c c c c}
    \toprule
    \multicolumn{5}{c}{\textbf{Non-IID Scenario (Dirichlet=0.3)}} \\
    \cmidrule(lr){1-5} 
    \textbf{R50-C100} & Class Label & \textbf{$A_{Te_{D_u}}{}_{\downarrow}$ } & \textbf{$A_{Te_{D_r}}{}_{\uparrow}$ } & MIA$_{\downarrow}$  \\
    \midrule
    Raw Model & - & - & 47.50\%& -\\
    Round 1& 1 & 0.0\%& 46.96\%& 4.67\%\\ 
    Round 2& 5 & 2.00\%& 46.14\%& 3.33\%\\ 
    Round 3& 10 & 0.0\%& 45.08\%& 3.67\%\\ 
    Round 4& 15 & 1.33\%& 44.68\%& 4.00\%\\ 
    Round 5 & 20 & 0.0\%& 44.07\%& 3.67\%\\ 
    \midrule
    \multicolumn{5}{c}{\textbf{Centralized Scenario}} \\
    \cmidrule(lr){1-5} 
    \textbf{R50-C100}& Class Label & \textbf{$A_{Te_{D_u}}{}_{\downarrow}$ } & \textbf{$A_{Te_{D_r}}{}_{\uparrow}$ } & MIA$_{\downarrow}$ \\
    \midrule
    Raw Model & - & - & 80.27\%& -\\
    Round 1 & 1 & 0.0\%& 79.70\%& 2.00\%\\ 
    Round 2 & 5 & 0.0\%& 79.23\%& 2.67\%\\ 
    Round 3 & 10 & 0.0\%& 78.20\%& 3.00\%\\ 
    Round 4 & 15 & 2.33\%& 77.48\%& 2.67\%\\ 
    Round 5 & 20 & 2.00\%& 77.19\%& 3.00\%\\ %
    \bottomrule
  \end{tabular}
  \label{tab:continuous_unlearning_comparison}
\end{table}

\noindent \textbf{Main Result}\ \ \ Table~\ref{tab:unlearning_iid_non_iid} presents the experimental results of client-level, class-level, and sample-level unlearning tasks under different client data distributions (with the best results in bold and the second-best underlined). In the class unlearning task, FedOrtho almost completely erases the knowledge of the target class while maintaining performance on the retained classes due to the orthogonal constraints and weight alignment mechanism. For client-level and sample-level unlearning, FedOrtho even outperforms the retrained model in unlearning effectiveness while sustaining the performance of the retained set. This advantage stems from the orthogonal constraints throughout training: they not only drive class-level semantic separation but also strengthen mutually exclusive feature representations at the sample level, thereby ensuring robust generalization under varying non-IID strengths.

On the retained set, FT achieves the optimal performance as it is retrained on the remaining data based on the original model. In contrast, FedOrtho leverages orthogonal constraints and OSP to accurately locate and prune target kernels. It recovers performance without any additional fine-tuning, ultimately matching FT’s performance. While the Fused method also delivers decent results, it relies on the retraining of sparse layers. Conversely, FedOrtho accomplishes unlearning through one-step pruning with no extra training procedures. It achieves the synergistic advantages of ``high retained accuracy and strong unlearning effectiveness and zero retraining overhead" across all tasks, highlighting its efficiency and stability under different data distributions.
The appendix~\ref{more_results} provides more comprehensive experimental results.

\noindent \textbf{Centralized Unlearning}\ \ \ Table~\ref{tab:unlearning_res50} presents the experimental results of unlearning in the centralized setting. Unlike existing methods that rely on backward gradient computation or generating adversarial examples, FedOrtho achieves effective unlearning solely through a single forward pass to calculate convolutional kernel activation differences, followed by a one-step pruning step. This design significantly reduces computational overhead, completing the unlearning operation in subseconds while achieving nearly complete knowledge erasure. As the second-best method, Salun outperforms traditional methods dependent on Fisher information or adversarial examples, it indeed accelerates unlearning speed and preserves decent retained performance but still cannot eliminate the need for additional fine-tuning. In contrast, FedOrtho requires no subsequent training procedures at all, offering superior efficiency and deployment flexibility.

\noindent \textbf{Continue Unlearning}\ \ \ Table~\ref{tab:continuous_unlearning_comparison} presents the results of five consecutive class unlearning steps under federated and centralized settings. The model loss induced by pruning is manageable, and the retained set performance exhibits stable and controllable slight degradation in both scenarios. Meanwhile, effective erasure of unlearned set knowledge and low privacy risks are achieved. Enabled by feature decoupling and orthogonal constraint mechanism, FedOrtho balances retained set stability and unlearning effectiveness without using raw data for model restoration.



\subsection{Ablation and Analysis}

We conduct the ablation to validate the impact of core design factors on unlearning and retained performance, including pruning ratio, pruning intensity, and the position of orthogonal constraints. Furthermore, we conduct the analysis of privacy and resource efficiency to validate an efficient and secure federated unlearning framework.


\begin{table}[t]
  \caption{Prune ratio \( r \) effect on class specific forgetting (for the first class) using ResNet-50/CIFAR-100.}
  \centering
  \fontsize{9pt}{10pt}\selectfont
  \begin{tabular}{c c c|c c}
    \toprule
    \multirow{2}{*}{\textbf{Prune Ratio \(r\)}} & \multicolumn{2}{c}{\textbf{Ortho.}} & \multicolumn{2}{c}{\textbf{Non-Ortho.}} \\
    \cmidrule(lr){2-3} \cmidrule(lr){4-5}
    & $A_{Te_{D_u}}{}_{\downarrow}$ & $A_{Te_{D_r}}{}_{\uparrow}$ & $A_{Te_{D_u}}{}_{\downarrow}$ & $A_{Te_{D_r}}{}_{\uparrow}$ \\
    \midrule
    Raw Model& 92.00\%& 80.15\%& 90.00\%&78.19\%\\ 
    0.005& 41.0\%& 80.13\%& 72.0\%& 77.21\%\\
    0.006& 16.0\%& 80.02\%& 64.0\%& 76.96\%\\
    0.007& 11.0\%& 79.99\%& 44.0\%& 76.43\%\\
    0.008& 9.0\%& 79.91\%& 38.0\%& 74.29\%\\
 0.009& 5.0\%& 79.82\%& 23.0\%&72.21\%\\
 0.01& 0.0\%& 79.70\%& 21.0\%&71.74\%\\
 \bottomrule
  \end{tabular}
  \label{tab:pruning}
\end{table}

\begin{table}[t]
\caption{Effect of pruning strength \(\alpha\) on class specific forgetting (for the first class) using ResNet-50/CIFAR-100}
  \centering
  \fontsize{9pt}{10pt}\selectfont
  \begin{tabular}{c cc}
    \toprule
    \textbf{Pruning Strength $\alpha$}& $A_{Te_{D_u}}{}_{\downarrow}$ & $A_{Te_{D_r}}{}_{\uparrow}$ \\
    \midrule
     Raw Model& 92.00\%&80.15\%\\
    1.0 & 0.0\%& 76.75\%\\
    0.9 & 0.0\%  & 77.00\%\\
    0.8 & 0.0\%  & 78.05\%\\
    0.7 & 0.0\%  & 78.14\%\\
    0.6 & 0.0\%  & 78.37\%\\
    \textbf{0.5} & \textbf{0.0\%}  & \textbf{79.70\%}  \\
    0.4 & 3.0\%& 79.95\%\\
    0.3 & 6.0\%& 80.10\%\\
    \bottomrule
  \end{tabular}
     \label{tab:strength}
\end{table}


\noindent \textbf{Pruning Ratio and Pruning Strength}\ \ \ Table~\ref{tab:pruning} compares the performance of orthogonal and non-orthogonal models under different pruning ratios. For a fixed ratio $r$, orthogonal models deliver more complete unlearning with minimal structural change while retaining high accuracy. On CIFAR and TinyImageNet, pruning as little as $r=0.01$ or $r=0.009$ fully removes the target, highlighting the model’s pruning sensitivity and validating the role of orthogonal constraints in feature separation and stable erasure.

Table~\ref{tab:strength} further illustrates the impact of pruning intensity \(\alpha\). By continuously scaling weights, soft pruning achieves a smoother unlearning-retention trade-off and effectively avoids feature collapse compared to hard zeroing pruning.

\begin{figure}[t]
    \centering
    \includegraphics[width=0.48\textwidth]{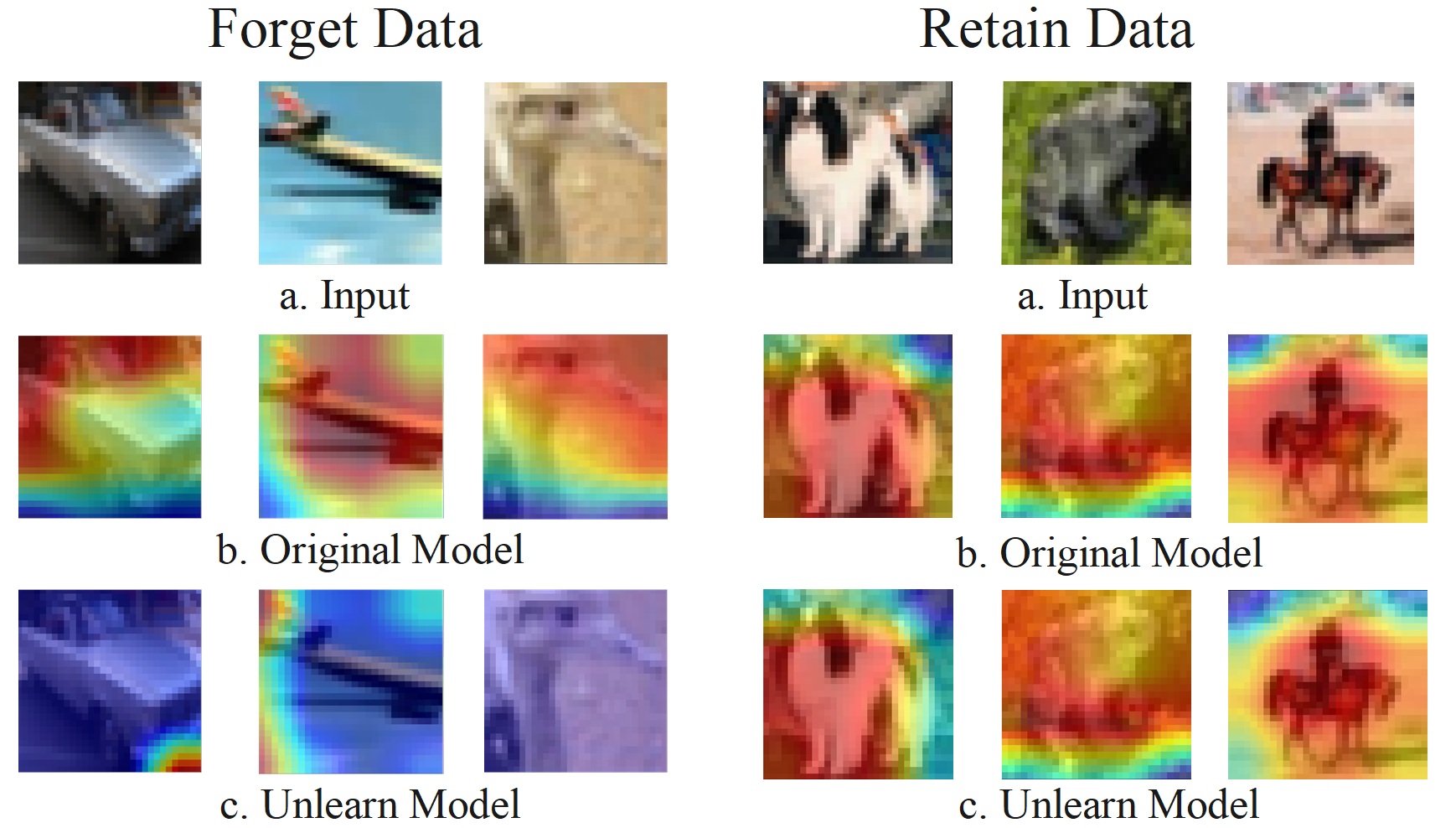}
    \caption{Grad-CAM heatmaps of CIFAR-10 before and after unlearning.}
    \label{fig:hotmap1.png}
\end{figure}

Grad-CAM results in Fig.~\ref{fig:hotmap1.png} show that the activation of the unlearned class is significantly weakened after unlearning, while the response of the retained classes remains nearly unchanged. This intuitively reflects the advantages of the orthogonal and soft pruning mechanisms in achieving semantic decoupling and stable retention. 
More case study are included in the appendix~\ref{sec:kernel_heatmap}.

\begin{table}[t]
\caption{Comparison of forgetting (for the first class) and retention performance of one-shot pruning for conv kernels at different layers with orthogonal constraints (ResNet-50/CIFAR-100).}
\centering
\fontsize{9pt}{10pt}\selectfont
\begin{tabular}{c c c c}
\toprule
\textbf{Orthogonal Layer} & $A_{\mathrm{Te}_{D_u}}{}_{\downarrow}$ & $A_{\mathrm{Te}_{D_r}}{}_{\uparrow}$  & Pruned Ratio \\
\midrule
Non-ortho & 21.00\%& 71.74\%&0.01\\
Shallow & 89.00\% & 79.00\% & 0.01 \\
Mid & 67.00\% & 78.17\% & 0.01 \\
\textbf{Deep} & \textbf{0.00\%} & \textbf{79.70\%} & \textbf{0.01} \\
All Layers & 0.00\% & 77.19\% & 0.01 \\
\bottomrule
\end{tabular}
\label{tab:layer_ortho}
\end{table}

\noindent \textbf{Orthogonal Constraints in Conv Layers}\ \ \ Table~\ref{tab:layer_ortho} indicates that deep-layer orthogonal constraints play a dominant role in unlearning: they promote feature decoupling in the semantic embedding space, yielding clear separation between the target and retained class distributions and enabling complete unlearning with a minimal pruning ratio. In contrast, shallow layer constraints only affect shared texture features and have a limited effect on semantic discrimination. More results in the Appendix~\ref{sec:motivational_observations} show that deep-layer decoupling is sufficient to support global feature separation.

\noindent \textbf{Privacy}\ \ \ From the MIA metric in Table~\ref{tab:unlearning_iid_non_iid}, FedOrtho's orthogonalization further prevents the reconstruction of pruned kernels, enhancing the irreversibility and privacy reliability of unlearning. In contrast, fine-tuning or retraining adjusts the model on retained data, creating additional privacy exposure. FedOrtho conducts one shot soft pruning using client supplied pruning masks without re-accessing retained data, enabling efficient unlearning while markedly reducing both communication cost and privacy leakage.

\begin{figure}[t]
    \centering
\includegraphics[width=1.0\linewidth]{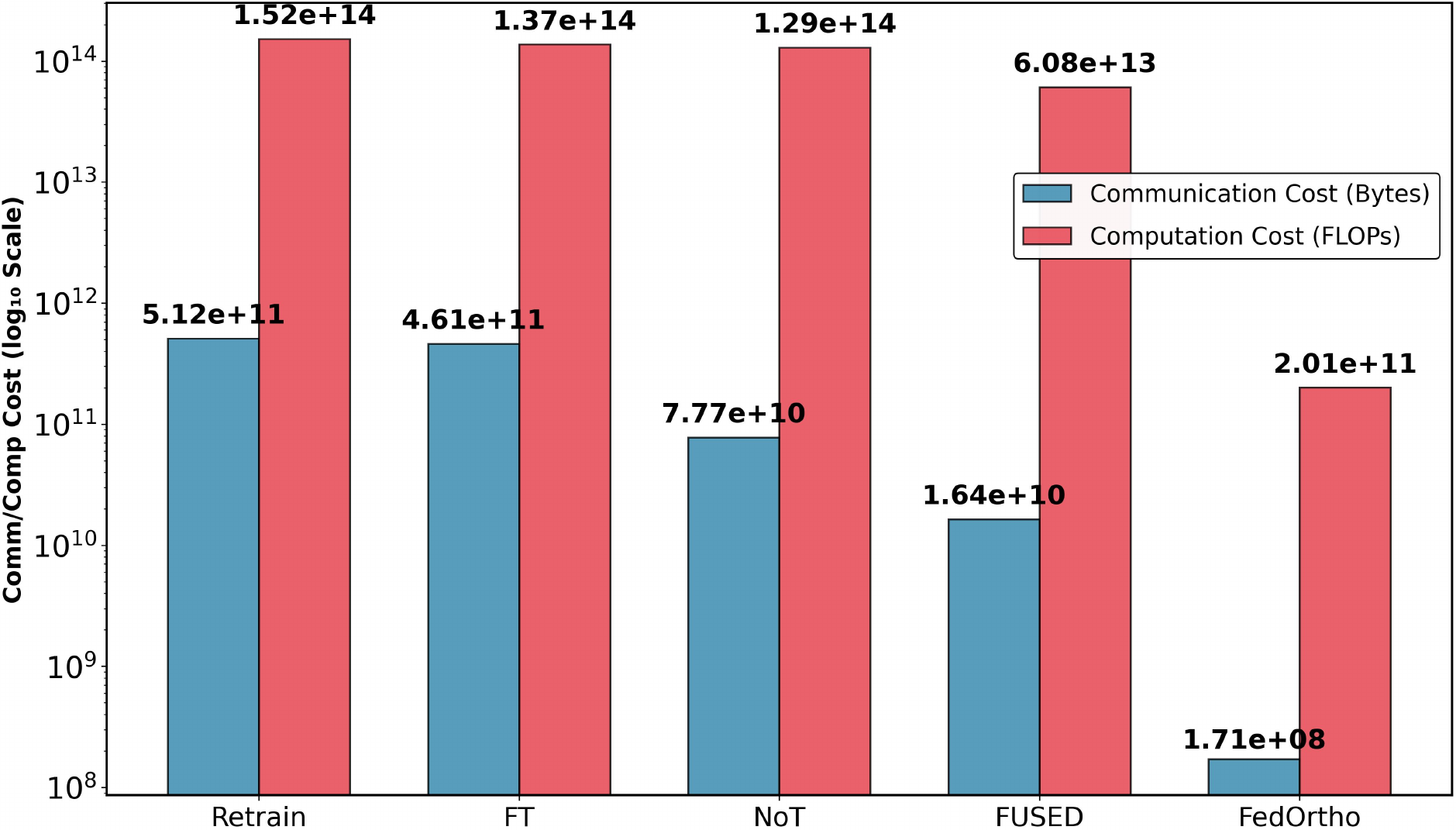}
    \caption{Communication and computation costs of federated unlearning process for different methods on ResNet-50/CIFAR-100.}
    \label{fig:placeholder}
\end{figure}

\noindent \textbf{Resource Efficiency}\ \ \ As shown in Fig.~\ref{fig:placeholder}, FedOrtho exhibits notable computational and communication advantages in federated unlearning scenarios. In terms of computation, the unlearning operation relies on one-shot soft pruning without requiring retraining, avoiding the gradient computation and parameter update overhead associated with fine-tuning or retraining. Notably, our method introduces orthogonality constraints during the initial training phase, which leads to a computational overhead of 1.32–1.45× compared to vanilla FL without such constraints. This upfront cost, however, is offset by the complete elimination of subsequent retraining processes. For communication, only conv kernel pruning masks from clients are needed, eliminating the need to transmit full model parameters or raw data. This reduces communication costs by at least 2-3 order of magnitude compared to baselines, enabling efficient and controllable knowledge unlearning.

\section{Limition}
Due to communication restrictions in FL, current experiments focus on CNNs and classification tasks, but the effectiveness of non-conv architectures (e.g., Transformers) remains to be verified. We plan to explore extending the orthogonal idea to deeper, multi-task networks in the future.

\section{Conclusion}
We propose FedOrtho, an efficient and controllable federated unlearning framework integrating deep layer conv kernel orthogonalization and adaptive soft pruning. Enforcing kernel orthogonality enables kernels to learn independent features, which establishes clear separability between forgotten and retained data for thorough erasure while maximizing retained performance. Extensive experiments across diverse datasets and architectures validate FedOrtho’s strong, sustainable unlearning capability and high efficiency.

\bibliographystyle{unsrt}  
\bibliography{ref}    
\clearpage
\setcounter{page}{1}

\section{Full Proof of Kernel Orthogonality-Induced Feature Decoupling}
\label{app:decoupling_proof}

\paragraph{Foundational Definition: Convolutional Functional in Dual Space}
The following definition formalizes the mathematical essence of convolutional responses, serving as the basis for subsequent derivations.

\begin{definition}[Convolutional Functional in Dual Space]
Let the input data space be $\mathbb{R}^D$, where $D = C_{\text{in}} \cdot k^2$ ($C_{\text{in}}$: number of input channels; $k$: convolutional kernel size). The linear functionals in its dual space $(\mathbb{R}^D)^*$ satisfy three properties:
\begin{enumerate}
    \item \textit{Unique Correspondence}: Each convolutional kernel $\boldsymbol{w}_j \in \mathbb{R}^D$ uniquely determines a functional $f_j \in (\mathbb{R}^D)^*$;
    \item \textit{Patch-Level Response}: For a flattened $k \times k$ input patch $\boldsymbol{x} \in \mathbb{R}^D$, the response of $f_j$ is 
    \begin{equation}
    f_j(\boldsymbol{x}) = \boldsymbol{w}_j^\top \boldsymbol{x},
    \label{eq:patch_response}
    \end{equation}
    \item \textit{Image-Level Response}: For an image $X$ partitioned into $m$ patches $\{\boldsymbol{x}_{u,v}\}_{1 \leq u,v \leq m}$, the model's feature response intensity to $X$ is the maximum of functional responses: 
    \begin{equation}
    \Phi_j(X) = \max_{1 \leq u,v \leq m} f_j(\boldsymbol{x}_{u,v}).
    \label{eq:image_response}
    \end{equation}
\end{enumerate}
\label{def:conv_dual_func}
\end{definition}

\paragraph{Step 1: Kernel Near-Orthogonality via Orthogonal Regularization}
We first introduce the orthogonal regularization constraint on convolutional kernels, then derive the near-orthogonality of kernel vectors (a critical precursor to functional irrelevance).

\begin{assumption}[Orthogonal Regularization Constraint]
Let $\boldsymbol{W} \in \mathbb{R}^{C_{\text{out}} \times D}$ denote the convolutional kernel matrix, where each row corresponds to a kernel vector $\boldsymbol{w}_j \in \mathbb{R}^D$ ($C_{\text{out}}$: number of output channels). The orthogonal regularization loss satisfies:
\begin{equation}
\mathcal{L}_{\text{ortho}} = \|\boldsymbol{W}\boldsymbol{W}^\top - \boldsymbol{I}\|_F \leq \varepsilon,
\label{eq:ortho_reg_loss}
\end{equation}
where $\boldsymbol{I}$ is the $C_{\text{out}} \times C_{\text{out}}$ identity matrix, and $\varepsilon > 0$ is the deviation from strict orthogonality.
\label{assump:ortho_constraint}
\end{assumption}

\begin{lemma}[Near-Orthogonality of Kernel Vectors]
If the orthogonal regularization constraint in Assumption~\ref{assump:ortho_constraint} holds, then:
\begin{enumerate}
    \item For any distinct kernel vectors $\boldsymbol{w}_i, \boldsymbol{w}_j$ ($i \neq j$): $|\boldsymbol{w}_i^\top \boldsymbol{w}_j| \leq \varepsilon$;
    \item For any kernel vector $\boldsymbol{w}_j$: $1 - \varepsilon \leq \|\boldsymbol{w}_j\|_2^2 \leq 1 + \varepsilon$.
\end{enumerate}
\label{lem:kernel_near_ortho}
\end{lemma}

\begin{proof}
By the definition of the Frobenius norm, the orthogonal regularization loss in Assumption~\ref{assump:ortho_constraint} expands to:
\begin{equation}
\|\boldsymbol{W}\boldsymbol{W}^\top - \boldsymbol{I}\|_F^2 = \sum_{i=1}^{C_{\text{out}}} \sum_{j=1}^{C_{\text{out}}} \left(\boldsymbol{w}_i^\top \boldsymbol{w}_j - \delta_{ij}\right)^2 \leq \varepsilon^2,
\label{eq:frobenius_expand}
\end{equation}
where $\delta_{ij}$ is the Kronecker delta (i.e., $\delta_{ij} = 1$ if $i = j$, and $\delta_{ij} = 0$ otherwise).

1. For $i \neq j$: The double sum reduces to a single term $(\boldsymbol{w}_i^\top \boldsymbol{w}_j)^2$. Since the sum of non-negative terms is bounded by $\varepsilon^2$, we have $(\boldsymbol{w}_i^\top \boldsymbol{w}_j)^2 \leq \varepsilon^2$. Taking the square root of both sides gives $|\boldsymbol{w}_i^\top \boldsymbol{w}_j| \leq \varepsilon$.

2. For $i = j$: The sum term becomes $(\|\boldsymbol{w}_j\|_2^2 - 1)^2$ (since $\boldsymbol{w}_j^\top \boldsymbol{w}_j = \|\boldsymbol{w}_j\|_2^2$ and $\delta_{jj} = 1$). Similarly, $(\|\boldsymbol{w}_j\|_2^2 - 1)^2 \leq \varepsilon^2$. Taking the square root and rearranging terms gives $1 - \varepsilon \leq \|\boldsymbol{w}_j\|_2^2 \leq 1 + \varepsilon$.
\end{proof}

\paragraph{Step 2: From Kernel Near-Orthogonality to Irrelevant Functional Responses}
To connect kernel orthogonality to functional behavior, we first formalize the statistical property of input patches, then derive that functional responses are statistically irrelevant via bounded covariance.

\begin{assumption}[Statistical Property of Input Patches]
Input patches $\boldsymbol{x} \in \mathbb{R}^D$ (after Batch Normalization) satisfy two conditions:
\begin{enumerate}
    \item \textit{Mean Zero}: $\mathbb{E}[\boldsymbol{x}] = 0$, where $\mathbb{E}[\cdot]$ denotes the expectation over the patch distribution;
    \item \textit{Approximate Whitening}: The covariance matrix of $\boldsymbol{x}$ is 
    \begin{equation}
    \Sigma = \mathbb{E}[\boldsymbol{x}\boldsymbol{x}^\top] = \boldsymbol{I} + \Delta,
    \label{eq:patch_covariance}
    \end{equation}
    where $\Delta$ is the whitening error matrix with spectral norm $\|\Delta\|_2 \leq \eta$ ($\eta > 0$ is the whitening error).
\end{enumerate}
\label{assump:patch_statistic}
\end{assumption}

\begin{theorem}[Near-Orthogonality of Functional Covariance]
For any two functionals $f_i, f_j \in (\mathbb{R}^D)^*$ (induced by kernel vectors $\boldsymbol{w}_i, \boldsymbol{w}_j$ via Definition~\ref{def:conv_dual_func}), their covariance satisfies:
\begin{equation}
|\text{Cov}(f_i, f_j)| \leq \varepsilon + (1 + \varepsilon)\eta =: \varepsilon_1,
\label{eq:func_cov_bound}
\end{equation}
where $\varepsilon_1$ is a small constant that approaches 0 as $\varepsilon \to 0$ (strict kernel orthogonality) and $\eta \to 0$ (perfect patch whitening).
\label{thm:func_cov_near_ortho}
\end{theorem}

\begin{proof}
By the definition of covariance, we have:
\begin{equation}
\text{Cov}(f_i, f_j) = \mathbb{E}[f_i(\boldsymbol{x})f_j(\boldsymbol{x})] - \mathbb{E}[f_i(\boldsymbol{x})]\mathbb{E}[f_j(\boldsymbol{x})].
\label{eq:covariance_def}
\end{equation}

From Assumption~\ref{assump:patch_statistic}, $\mathbb{E}[\boldsymbol{x}] = 0$. For any functional $f_k(\boldsymbol{x}) = \boldsymbol{w}_k^\top \boldsymbol{x}$ (via Definition~\ref{def:conv_dual_func}), this implies $\mathbb{E}[f_k(\boldsymbol{x})] = \boldsymbol{w}_k^\top \mathbb{E}[\boldsymbol{x}] = 0$. Thus, the covariance simplifies to:
\begin{equation}
\text{Cov}(f_i, f_j) = \mathbb{E}[f_i(\boldsymbol{x})f_j(\boldsymbol{x})] = \mathbb{E}[\boldsymbol{w}_i^\top \boldsymbol{x} \cdot \boldsymbol{w}_j^\top \boldsymbol{x}].
\label{eq:covariance_simplify}
\end{equation}

Using the linearity of expectation and the definition of covariance matrix $\Sigma = \mathbb{E}[\boldsymbol{x}\boldsymbol{x}^\top]$ (from Assumption~\ref{assump:patch_statistic}), we rewrite the expectation as:
\begin{equation}
\mathbb{E}[\boldsymbol{w}_i^\top \boldsymbol{x} \cdot \boldsymbol{w}_j^\top \boldsymbol{x}] = \boldsymbol{w}_i^\top \mathbb{E}[\boldsymbol{x}\boldsymbol{x}^\top] \boldsymbol{w}_j = \boldsymbol{w}_i^\top \Sigma \boldsymbol{w}_j.
\label{eq:expectation_rewrite}
\end{equation}

Substitute $\Sigma = \boldsymbol{I} + \Delta$ (from Assumption~\ref{assump:patch_statistic}) and split the term:
\begin{equation}
\boldsymbol{w}_i^\top \Sigma \boldsymbol{w}_j = \boldsymbol{w}_i^\top \boldsymbol{w}_j + \boldsymbol{w}_i^\top \Delta \boldsymbol{w}_j.
\label{eq:sigma_split}
\end{equation}

By the triangle inequality, $|\boldsymbol{a} + \boldsymbol{b}| \leq |\boldsymbol{a}| + |\boldsymbol{b}|$, so:
\begin{equation}
|\boldsymbol{w}_i^\top \Sigma \boldsymbol{w}_j| \leq |\boldsymbol{w}_i^\top \boldsymbol{w}_j| + |\boldsymbol{w}_i^\top \Delta \boldsymbol{w}_j|.
\label{eq:triangle_inequality}
\end{equation}

For the second term, apply the Cauchy-Schwarz inequality: $|\boldsymbol{a}^\top \Delta \boldsymbol{b}| \leq \|\boldsymbol{a}\|_2 \|\Delta\|_2 \|\boldsymbol{b}\|_2$. From Lemma~\ref{lem:kernel_near_ortho}, $\|\boldsymbol{w}_i\|_2 \leq \sqrt{1 + \varepsilon}$ and $\|\boldsymbol{w}_j\|_2 \leq \sqrt{1 + \varepsilon}$; from Assumption~\ref{assump:patch_statistic}, $\|\Delta\|_2 \leq \eta$. Thus:
\begin{equation}
|\boldsymbol{w}_i^\top \Delta \boldsymbol{w}_j| \leq \sqrt{1 + \varepsilon} \cdot \eta \cdot \sqrt{1 + \varepsilon} = (1 + \varepsilon)\eta.
\label{eq:cauchy_schwarz_bound}
\end{equation}

From Lemma~\ref{lem:kernel_near_ortho}, $|\boldsymbol{w}_i^\top \boldsymbol{w}_j| \leq \varepsilon$. Substituting both bounds into Eq.\eqref{eq:triangle_inequality} gives Eq.\eqref{eq:func_cov_bound}.
\end{proof}

\paragraph{Step 3: From Irrelevant Functionals to Feature Decoupling}
We first define class-specific functionals and additional statistical assumptions, then derive that irrelevant functionals lead to complete feature decoupling between target and retained classes.

\begin{definition}[Class-Specific Functional Sets]
Let $A_{\text{target},j} = \mathbb{E}_{X \sim X_{\text{target}}}[\Phi_j(X)]$ (average image-level response of $f_j$ on target-class images $X_{\text{target}}$) and $A_{\text{retain},j} = \mathbb{E}_{X \sim X_{\text{retain}}}[\Phi_j(X)]$ (average response on retained-class images $X_{\text{retain}}$). For a threshold $\tau > 0$:
\begin{enumerate}
    \item \textit{Target-Data Functional Set}: $\mathcal{F}_{\text{target}} = \{f_j \in (\mathbb{R}^D)^* \mid A_{\text{target},j} - A_{\text{retain},j} > \tau\}$ (functionals strongly associated with the target data);
    \item \textit{Retained-Data Functional Set}: $\mathcal{F}_{\text{retain}} = (\mathbb{R}^D)^* \setminus \mathcal{F}_{\text{target}}$ (functionals associated with retained data, complementary to $\mathcal{F}_{\text{target}}$).
\end{enumerate}
Let $m_{\text{eff}}$ denote the number of \textit{effectively independent patches} in $X$ (correcting for correlation induced by patch overlap or stride).
\label{def:class_specific_func}
\end{definition}

\begin{assumption}[Sub-Gaussian Property of Functionals]
Each functional $f_j(\boldsymbol{x})$ (induced via Definition~\ref{def:conv_dual_func}) is a sub-Gaussian random variable: there exists a constant $\sigma > 0$ such that for all $t \in \mathbb{R}$,
\begin{equation}
\mathbb{E}\left[e^{t f_j(\boldsymbol{x})}\right] \leq e^{\sigma^2 t^2 / 2},
\label{eq:subgaussian_property}
\end{equation}
and the variance of $f_j(\boldsymbol{x})$ satisfies $\text{Var}(f_j(\boldsymbol{x})) \leq \sigma^2$.
\label{assump:func_subgaussian}
\end{assumption}

\begin{assumption}[Strong Activation of Target-Data Functionals]
There exists a constant $\gamma > 0$ such that for all $f_i \in \mathcal{F}_{\text{target}}$ (via Definition~\ref{def:class_specific_func}), the average response on target-data images satisfies:
\begin{equation}
A_{\text{target},i} \geq \gamma.
\label{eq:target_activation_lower}
\end{equation}
\label{assump:target_strong_activation}
\end{assumption}

\begin{theorem}[Data-Specific Role Separation]
For any $f_i \in \mathcal{F}_{\text{target}}$ (via Definition~\ref{def:class_specific_func}) and $f_j \in \mathcal{F}_{\text{retain}}$ (via Definition~\ref{def:class_specific_func}), the average response of $f_j$ on target-class images satisfies:
\begin{equation}
A_{\text{target},j} \leq C \left( \varepsilon_1 + \sigma\sqrt{\frac{\log m_{\text{eff}}}{m_{\text{eff}}}} \right) =: \epsilon_2,
\label{eq:class_response_bound}
\end{equation}
where $C > 0$ is a model-agnostic constant, and $\epsilon_2 \to 0$ as $\varepsilon_1 \to 0$ (via Theorem~\ref{thm:func_cov_near_ortho}) and $m_{\text{eff}} \to \infty$ (sufficient independent patches).
\label{thm:class_role_separation}
\end{theorem}

\begin{proof}
The proof proceeds in three steps, combining conditional expectation control and extreme value bounds for sub-Gaussian variables.

\paragraph{1: Bounded Conditional Expectation}
For a patch $\boldsymbol{x}$, let $f_i(\boldsymbol{x}) = t$ (response of the target-class functional $f_i$). Since $f_i$ and $f_j$ are linear functionals of $\boldsymbol{x}$ (via Definition~\ref{def:conv_dual_func}), their conditional expectation satisfies a linear relationship:
\begin{equation}
\mathbb{E}[f_j(\boldsymbol{x}) \mid f_i(\boldsymbol{x}) = t] = \frac{\text{Cov}(f_i, f_j)}{\text{Var}(f_i)} \cdot t.
\label{eq:conditional_expectation}
\end{equation}

From Theorem~\ref{thm:func_cov_near_ortho}, $|\text{Cov}(f_i, f_j)| \leq \varepsilon_1$; from Assumption~\ref{assump:func_subgaussian}, $\text{Var}(f_i) \geq \sigma^2$ (lower bound of sub-Gaussian variance). Substituting these bounds gives:
\begin{equation}
|\mathbb{E}[f_j(\boldsymbol{x}) \mid f_i(\boldsymbol{x}) = t]| \leq \frac{\varepsilon_1}{\sigma^2} \cdot |t|.
\label{eq:conditional_bound}
\end{equation}

From Assumption~\ref{assump:target_strong_activation}, $A_{\text{target},i} \geq \gamma$, but $\varepsilon_1$ is a small constant (via Theorem~\ref{thm:func_cov_near_ortho}), so the conditional expectation of $f_j$ is bounded by $O(\varepsilon_1)$ (local response suppression).

\paragraph{2: Extreme Value Bound for Sub-Gaussian Variables}
For $m_{\text{eff}}$ effectively independent patches, the responses $\{f_j(\boldsymbol{x}_{u,v})\}_{1 \leq u,v \leq m_{\text{eff}}}$ are independent sub-Gaussian variables (via Assumption~\ref{assump:func_subgaussian}). By the Borell-TIS inequality (a fundamental result for sub-Gaussian extremes), the expectation of their maximum satisfies:
\begin{equation}
\mathbb{E}\left[\max_{1 \leq u,v \leq m_{\text{eff}}} f_j(\boldsymbol{x}_{u,v})\right] \leq O\left( \sigma\sqrt{\frac{\log m_{\text{eff}}}{m_{\text{eff}}}} \right).
\label{eq:subgaussian_extreme_bound}
\end{equation}

This bound shows that the global maximum response of $f_j$ decays to 0 as $m_{\text{eff}}$ increases (global response suppression).

\paragraph{3: Merge Local and Global Bounds}
The average response $A_{\text{target},j} = \mathbb{E}_{X \sim X_{\text{target}}}[\Phi_j(X)]$ (via Definition~\ref{def:class_specific_func}) is the expectation of the maximum functional response over patches. Combining the local conditional expectation bound Eq.\eqref{eq:conditional_bound} and global extreme value bound Eq.\eqref{eq:subgaussian_extreme_bound}, we obtain Eq.\eqref{eq:class_response_bound}, where $C > 0$ is a constant integrating the two bounds.
\end{proof}

\begin{theorem}[Feature Decoupling]
For any $f_i \in \mathcal{F}_{\text{target}}$ (via Definition~\ref{def:class_specific_func}) and $f_j \in \mathcal{F}_{\text{retain}}$ (via Definition~\ref{def:class_specific_func}), the covariance of their image-level responses on target-data images satisfies:
\begin{equation}
\mathbb{E}[\Phi_i(X) \Phi_j(X)] \leq B \cdot \epsilon_2 =: \delta,
\label{eq:feature_decoupling_final}
\end{equation}
where $B > 0$ is the upper bound of $\Phi_i(X)$ (activation saturation in deep networks), and $\delta \to 0$ as $\varepsilon \to 0$, $\eta \to 0$, and $m_{\text{eff}} \to \infty$.
\label{thm:feature_decoupling}
\end{theorem}

\begin{proof}
By the law of total expectation, the expectation of the product $\Phi_i(X) \Phi_j(X)$ can be bounded using the maximum response of $\Phi_i(X)$:
\begin{equation}
\mathbb{E}[\Phi_i(X) \Phi_j(X)] \leq \max_{X \sim X_{\text{target}}} \Phi_i(X) \cdot \mathbb{E}[\Phi_j(X)].
\label{eq:product_expectation_bound}
\end{equation}

Let $B = \max_{X \sim X_{\text{target}}} \Phi_i(X)$. From Definition~\ref{def:class_specific_func}, $\mathbb{E}[\Phi_j(X)] = A_{\text{target},j}$, so:
\begin{equation}
\mathbb{E}[\Phi_i(X) \Phi_j(X)] \leq B \cdot A_{\text{target},j}.
\label{eq:activation_saturation_bound}
\end{equation}

From Theorem~\ref{thm:class_role_separation}, $A_{\text{target},j} \leq \epsilon_2$. Substituting this bound into Eq.\eqref{eq:activation_saturation_bound} gives Eq.\eqref{eq:feature_decoupling_final}.

To confirm $\delta \to 0$, we trace the error chain:
 \begin{enumerate}
     \item From Theorem~\ref{thm:func_cov_near_ortho}, $\varepsilon \to 0$ and $\eta \to 0$ imply $\varepsilon_1 \to 0$ (via Eq.\eqref{eq:func_cov_bound});
    \item From Theorem~\ref{thm:class_role_separation}, $\varepsilon_1 \to 0$ and $m_{\text{eff}} \to \infty$ imply $\epsilon_2 \to 0$ (via Eq.\eqref{eq:class_response_bound});
    \item $\delta = B \cdot \epsilon_2 \to 0$, meaning the responses of target-data and retained-data functionals are asymptotically uncorrelated, which means feature decoupling is achieved.
 \end{enumerate}
\end{proof}

\section{Motivational Observations}
\label{sec:motivational_observations}

\subsection{Feature Similarity and Redundancy Analysis}
\label{sec:feature_similarity}

\begin{figure}[t]
    \centering
    \includegraphics[width=0.48\textwidth]{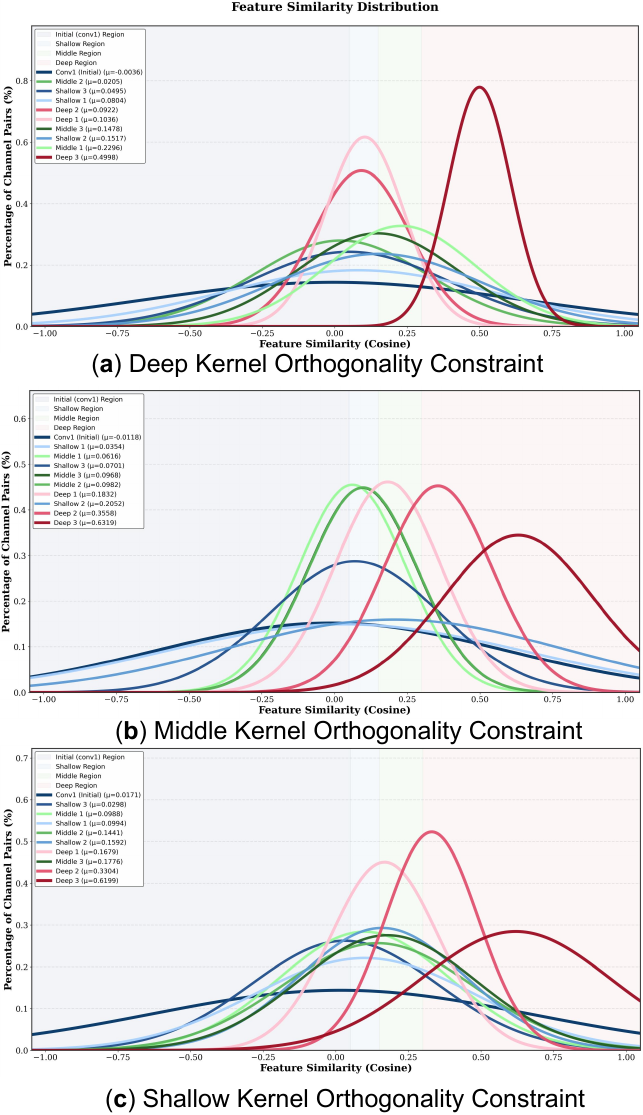}
    \caption{Feature Similarity Visualization and Redundancy Suppression via Orthogonality Constraints on Shallow, Middle, and Deep Layers}
    \label{fig:supply_similarity}
\end{figure}

Fig.~\ref{fig:supply_similarity} presents the feature similarity distribution curves of convolutional kernels in the shallow, middle, and deep layers after applying orthogonality constraints (the x-axis represents feature similarity values, and the y-axis denotes the proportion of samples with the corresponding similarity). We select 3 key convolutional layers from each of the three hierarchical levels (shallow, middle, and deep) to intuitively illustrate the statistical distribution characteristics of pairwise feature similarity through the distribution curves. Experimental results show that: when orthogonality constraints are only applied to deep convolutional kernels (Fig.~\ref{fig:supply_similarity} (a)), the peak of the curve shifts significantly toward the low-similarity interval, and the proportion of low-similarity samples increases substantially, indicating that redundant correlations among deep features are effectively weakened. In contrast, when orthogonality constraints are applied to shallow or middle layers (Fig.~\ref{fig:supply_similarity} (b)-(c)), the similarity distribution curve of deep features remains concentrated in the high-similarity interval, with no significant increase in the proportion of low-similarity samples, making it difficult to achieve efficient suppression of deep feature redundancy. This result fully verifies that orthogonality constraints on deep convolutional kernels are a key means to guide deep features toward a low-similarity distribution and reduce the redundancy of feature representations.

\subsection{Gram Matrix Visualization Analysis}
\begin{figure*}[t]
    \centering
    \includegraphics[width=1.0\textwidth]{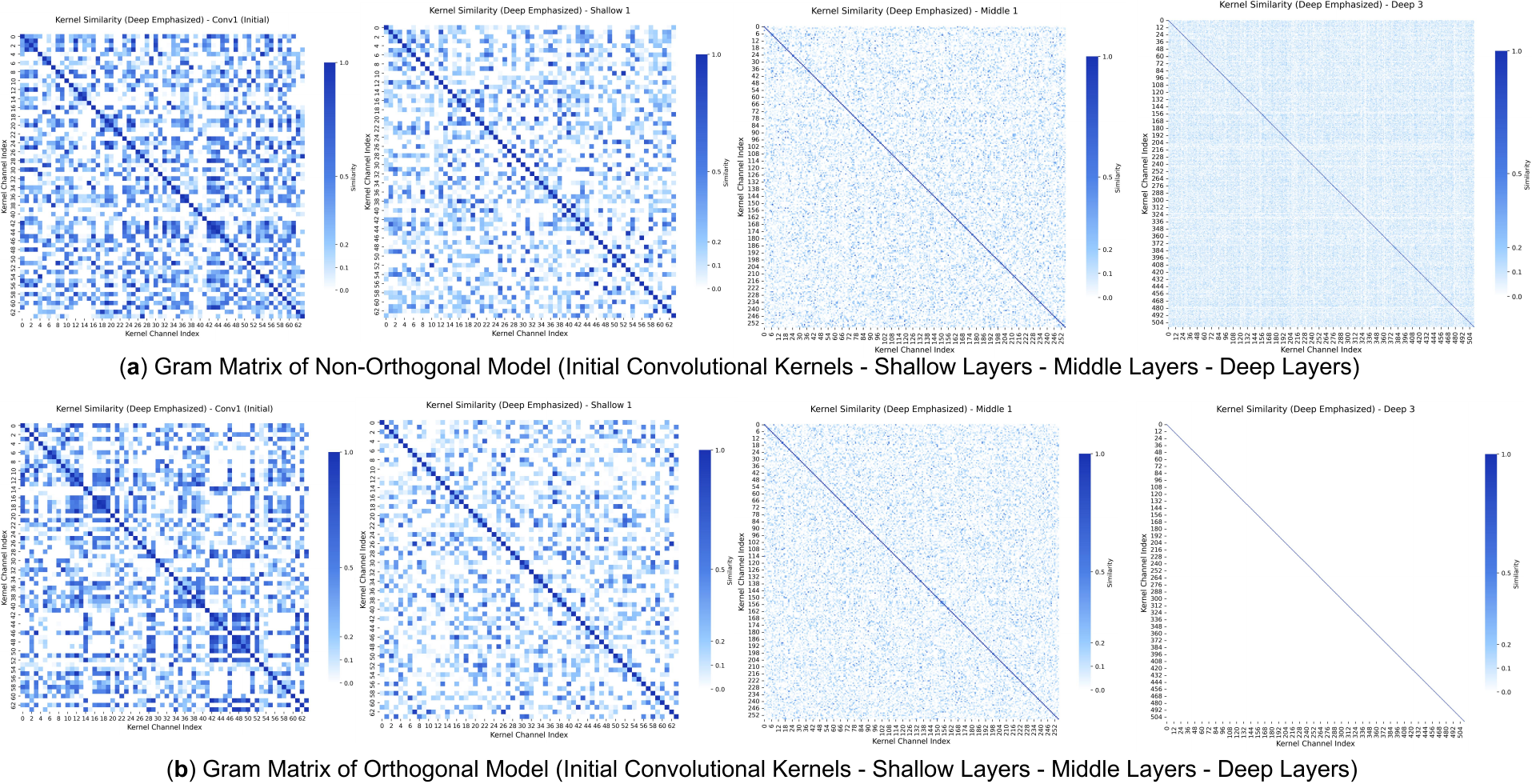}
    \caption{Gram matrix comparison across layers, showing deep-layer orthogonalization reduces redundancy while preserving shallow features.}
    \label{fig:supply_gram}
\end{figure*}

Fig.~\ref{fig:supply_gram} presents the hierarchical Gram matrix heatmaps of the baseline non-orthogonal model and the model with orthogonality constraints only applied to deep convolutional kernels, covering the initial convolutional kernels and three key convolutional layers (shallow, middle, and deep). As a core quantitative indicator of orthogonality in the feature space, the element $G_{i,j}$ of the Gram matrix essentially corresponds to the inner product of the $i$-th and $j$-th convolutional kernel feature maps. Its magnitude directly reflects the linear correlation between cross-channel feature maps—when the inner product is close to zero, the feature maps are approximately orthogonal in the Hilbert space, achieving the lowest feature redundancy. Notably, comparing the Gram matrix distributions of the shallow layers and initial convolutional kernels in Fig.~\ref{fig:supply_gram}(a) and Fig.~\ref{fig:supply_gram}(b), there is no significant difference in the off-diagonal response intensity and diagonal energy distribution between the two. This indicates that the orthogonality constraint has strict layer-specificity in optimizing deep features, without interfering with the inherent representational ability of shallow and initial convolutional kernels for basic visual features (e.g., edges, textures), thereby preserving the native representational bias of the network. Combined with the overall performance metrics, this layer-specific constraint strategy not only effectively suppresses the redundancy of deep high-level semantic features but also maintains the integrity of shallow basic features. It fully verifies the method’s favorable trade-off between feature decoupling and performance preservation, while being highly consistent with the functional division of labor in deep networks, where shallow layers extract basic features and deep layers model high-level semantics.

\section{Experiment Settings}
\label{exp_setting}
\subsection{Federated Training Setup}
Federated learning training is conducted on a hardware platform equipped with 1 NVIDIA RTX 4090 GPU (24GB VRAM), paired with Intel Xeon Platinum 24-core processors, 256GB DDR5 memory, and a 4TB NVMe SSD. It runs on the Ubuntu 22.04 system with Python 3.9 and the PyTorch 2.1.0 framework (supported by CUDA 12.1). The experiments use the CIFAR-100 (100 classes, 32×32) and TinyImageNet (200 classes, 64×64) datasets, deploying 100 virtual clients. Data is partitioned following a Dirichlet distribution (\(\alpha=0.3/0.6\)to control non-IID distribution), with 10\% of clients randomly selected per round for training. The models used are ResNet-18, ResNet-50, and VGG-16 (all adapted to input sizes and class numbers). Training employs the SGD optimizer (momentum 0.9, weight decay \(1 \times 10^{-3}\)) with an initial learning rate of 0.1, combined with a per-round learning rate decay of 0.998. A total of 1000 global communication rounds are performed, with 5 local epochs per round and a batch size of 50 (using mixed-precision training). Model aggregation adopts the FedAvg strategy, with orthogonality constraints (\(\lambda_{\text{ortho}}=0.1\), parameter search range: 0.01, 0.05, 0.1, 0.2, 0.5) and regularization (\(\lambda_{\text{reg}}=0.1\), parameter search range: 0.01, 0.05, 0.1, 0.2, 0.5) applied to deep convolutional layers. Data is augmented via random cropping, horizontal flipping, and random erasing, normalized using dataset-specific statistics. For the unlearning phase, fine-tuning is performed on all non-forgetting clients with a learning rate of 0.001. Results are reported as the mean over 3 runs with different random seeds (0, 20, 23).

\subsection{Centralized Training Setup}
Experiments are conducted on a workstation equipped with dual 14-core Intel Xeon CPUs and an NVIDIA TITAN Xp GPU (12GB VRAM), running on the Ubuntu 20.04 operating system with CUDA 11.7 and PyTorch 1.13 as the deep learning framework. The training uses the SGD optimizer with a momentum of 0.9 and a weight decay of $5 \times 10^{-4}$, with an initial learning rate of 0.1. The batch size is set to 256 for TinyImageNet and 128 for CIFAR-10/100. The training is performed for 200 epochs on TinyImageNet with a multi-step learning rate scheduler (milestones at 30 and 60 epochs) and 150 epochs on CIFAR datasets with cosine annealing. Models are adapted to match the input resolution and label dimension of each dataset. For data preprocessing, all images are normalized using dataset-specific statistics and augmented with random cropping and horizontal flipping. Experimental results are averaged over three independent runs with different random seeds to ensure reliability.

\section{Unlearning Performance Under Different Settings}
\label{more_results}
We conducted federated unlearning and centralized class-level unlearning experiments under different scenarios to verify the effectiveness of our proposed method. In the federated learning scenario, based on the ResNet18 model, we completed three types of unlearning tasks (client unlearning, class unlearning, and sample unlearning) under IID and Non-IID data distributions ($\alpha$=0.6 for weak heterogeneity and $\alpha$=0.3 for strong heterogeneity). By comparing with classical methods such as Retrain, FT, NoT, and FUSED, we verified the comprehensive advantages of our proposed method (FedOrtho) from three dimensions: forgetting accuracy ($A_{\mathrm{Te}_{D_u}}$), retention accuracy ($A_{\mathrm{Te}_{D_r}}$), and privacy protection metric (MIA). In the centralized class-level unlearning scenario, three models (Res18, Res34, and Vgg16) were selected, and comparisons were made with mainstream methods such as Retrain, SSD, EMNI, and TF-IDF on three major datasets (CIFAR-10, CIFAR-100, and Tiny ImageNet). In addition to the aforementioned metrics, training time evaluation was added. The results show that our proposed method not only ensures forgetting effectiveness and retention performance but also has extremely short unlearning time, achieving excellent comprehensive performance.


\subsection{Federated Unlearning Experiments}
We conducted federated unlearning experiments under different scenarios to verify the effectiveness of our proposed method. Based on the ResNet18 model, we completed three types of unlearning tasks (client unlearning, class unlearning, and sample unlearning) under IID and Non-IID data distributions ($\alpha$=0.6 for weak heterogeneity and \(\alpha\)=0.3 for strong heterogeneity). By comparing with classical methods such as Retrain, FT, NoT, and FUSED, we verified the comprehensive advantages of our proposed method (FedOrtho) from three dimensions: forgetting accuracy ($A_{\mathrm{Te}_{D_u}}$), retention accuracy ($A_{\mathrm{Te}_{D_r}}$), and privacy protection metric (MIA).

\begin{table*}[t]
\caption{Federated unlearning performance of ResNet18 on IID/Non-IID datasets. Metrics:\(A_{D_u}\)/\(A_{\mathrm{Te}_{D_u}}\) (Forget Accuracy on revoked training/test data),\(A_{\mathrm{Te}_{D_r}}\) (Retain Accuracy), MIA (privacy metric).}
    \centering
    \resizebox{\linewidth}{!}{
    \begin{tabular}{c c ccc | ccc | ccc}
        \toprule
        \multicolumn{11}{c}{\textbf{Non-IID (Dirichlet=0.6)}} \\ 
        \midrule
        \multirow{2}{*}{\textbf{Model}} & \multirow{2}{*}{\textbf{Method}} 
        & \multicolumn{3}{c}{\textbf{Client Unlearning}} 
        & \multicolumn{3}{c}{\textbf{Class Unlearning}} 
        & \multicolumn{3}{c}{\textbf{Sample Unlearning}} \\ 
        \cmidrule(lr){3-5}  \cmidrule(lr){6-8}  \cmidrule(lr){9-11} 
        & & $A{_{D_u}} \downarrow$& $A_{\mathrm{Te}_{D_r}} \uparrow$ & $\mathrm{MIA} \downarrow$ 
        & $A_{\mathrm{Te}_{D_u}} \downarrow$ & $A_{\mathrm{Te}_{D_r}} \uparrow$ & $\mathrm{MIA} \downarrow$ 
        & $A_{{D_u}} \downarrow$& $A_{\mathrm{Te}_{D_r}} \uparrow$ & $\mathrm{MIA} \downarrow$ \\ 
        \midrule

        \multirow{5}{*}{ResNet18} 
        & Retrain & 36.80±\footnotesize{0.92} & 46.72±\footnotesize{0.23} & 89.00±\footnotesize{0.58} & 0.00±\footnotesize{0.00} & 46.91±\footnotesize{0.45} & 75.93±\footnotesize{0.00}& 41.00±\footnotesize{1.05} & 46.57±\footnotesize{0.19} & 94.82±\footnotesize{0.47} \\ 
        & FT      & 40.00±\footnotesize{1.15} & \textbf{46.87}±\footnotesize{0.62} & 92.80±\footnotesize{0.70} & 4.93±\footnotesize{0.44} & \textbf{47.45}±\footnotesize{0.17} & 87.33±\footnotesize{0.93} & 44.00±\footnotesize{0.88} & \textbf{47.89}±\footnotesize{0.03} & 98.67±\footnotesize{0.35} \\
        & NoT     & \underline{37.00}±\footnotesize{2.98} & 44.65±\footnotesize{0.66} & 87.46±\footnotesize{1.95} & 7.26±\footnotesize{0.53} & 46.78±\footnotesize{0.91} & \underline{72.17}±\footnotesize{0.29} & \underline{38.33}±\footnotesize{1.17} & 47.12±\footnotesize{0.96} & 94.33±\footnotesize{0.54} \\
        & FUSED   & 35.73±\footnotesize{1.03} & 39.41±\footnotesize{0.85} & \underline{67.87}±\footnotesize{0.67} & 6.00±\footnotesize{0.22} & 41.57±\footnotesize{1.07} & 57.43±\footnotesize{0.23} & 35.73±\footnotesize{0.67}& 40.58±\footnotesize{1.13} & \underline{72.67}±\footnotesize{0.71} \\
        & FedOrtho    & \cellcolor{gray!20}\textbf{33.06}±\footnotesize{0.04} & \cellcolor{gray!20}\underline{44.92}±\footnotesize{0.25} & \cellcolor{gray!20}\textbf{57.33}±\footnotesize{0.55} & \cellcolor{gray!20}\textbf{0.00}±\footnotesize{0.00} & \cellcolor{gray!20}\underline{46.13}±\footnotesize{0.06} & \cellcolor{gray!20}\textbf{51.13}±\footnotesize{0.26} & \cellcolor{gray!20}\textbf{36.87}±\footnotesize{0.08} & \cellcolor{gray!20}\underline{46.31}±\footnotesize{0.21} & \cellcolor{gray!20}\textbf{65.33}±\footnotesize{0.73}\\ 
        \midrule

        \multicolumn{11}{c}{\textbf{Non-IID (Dirichlet=0.3)}} \\ 
        \midrule
        \multirow{2}{*}{\textbf{Model}} & \multirow{2}{*}{\textbf{Method}} 
        & \multicolumn{3}{c}{\textbf{Client Unlearning}} 
        & \multicolumn{3}{c}{\textbf{Class Unlearning}} 
        & \multicolumn{3}{c}{\textbf{Sample Unlearning}} \\ 
        \cmidrule(lr){3-5}  \cmidrule(lr){6-8}  \cmidrule(lr){9-11} 
        & & $A_{{D_u}} \downarrow$& $A_{\mathrm{Te}_{D_r}} \uparrow$ & $\mathrm{MIA} \downarrow$ 
        & $A_{\mathrm{Te}_{D_u}} \downarrow$ & $A_{\mathrm{Te}_{D_r}} \uparrow$ & $\mathrm{MIA} \downarrow$ 
        & $A_{{D_u}} \downarrow$& $A_{\mathrm{Te}_{D_r}} \uparrow$ & $\mathrm{MIA} \downarrow$ \\ 
        \midrule

        \multirow{5}{*}{ResNet18} 
        & Retrain & 33.20±\footnotesize{0.93} & 43.88±\footnotesize{0.29} & 90.47±\footnotesize{0.66} & 0.00±\footnotesize{0.00} & 44.35±\footnotesize{0.91} & 77.26±\footnotesize{0.00}& 38.67±\footnotesize{1.04} & 43.86±\footnotesize{0.15} & 96.67±\footnotesize{0.45} \\ 
        & FT      & 42.68±\footnotesize{1.10} & \textbf{44.52}±\footnotesize{0.94} & 94.32±\footnotesize{0.60} & 6.75±\footnotesize{0.49} & \textbf{45.19}±\footnotesize{0.71} & 89.14±\footnotesize{0.95} & 43.59±\footnotesize{1.07} & \textbf{45.37}±\footnotesize{0.20} & 99.21±\footnotesize{0.34} \\
        & NoT     & \underline{39.73}±\footnotesize{2.89} & \underline{42.03}±\footnotesize{0.04} & 89.20±\footnotesize{0.68} & 8.93±\footnotesize{0.54} & 43.96±\footnotesize{0.87} & \underline{75.76}±\footnotesize{2.38} & \underline{39.67}±\footnotesize{1.05} & \underline{42.97}±\footnotesize{0.90} & 96.00±\footnotesize{0.47} \\
        & FUSED   & 38.13±\footnotesize{0.99} & 39.29±\footnotesize{0.89} & \underline{69.47}±\footnotesize{0.65} & 8.00±\footnotesize{0.24} & 37.84±\footnotesize{0.92} & 59.20±\footnotesize{0.30} & 34.00±\footnotesize{0.98} & 41.42±\footnotesize{0.88} & \underline{75.00}±\footnotesize{0.64} \\
        & FedOrtho    & \cellcolor{gray!20}\textbf{31.06}±\footnotesize{0.05} & \cellcolor{gray!20}41.57±\footnotesize{0.11} & \cellcolor{gray!20}\textbf{69.80}±\footnotesize{0.73} & \cellcolor{gray!20}\textbf{0.00}±\footnotesize{0.00} & \cellcolor{gray!20}\underline{43.62}±\footnotesize{0.04} & \cellcolor{gray!20}\textbf{63.33}±\footnotesize{0.32} & \cellcolor{gray!20}\textbf{37.00}±\footnotesize{0.03} & \cellcolor{gray!20}42.79±\footnotesize{0.17} & \cellcolor{gray!20}\textbf{67.58}±\footnotesize{0.52}\\ 
        \midrule

        \multicolumn{11}{c}{\textbf{IID}} \\ 
        \midrule
        \multirow{2}{*}{\textbf{Model}} & \multirow{2}{*}{\textbf{Method}} 
        & \multicolumn{3}{c}{\textbf{Client Unlearning}} 
        & \multicolumn{3}{c}{\textbf{Class Unlearning}} 
        & \multicolumn{3}{c}{\textbf{Sample Unlearning}} \\ 
        \cmidrule(lr){3-5}  \cmidrule(lr){6-8}  \cmidrule(lr){9-11} 
        & & $A_{{D_u}} \downarrow$& $A_{\mathrm{Te}_{D_r}} \uparrow$ & $\mathrm{MIA} \downarrow$ 
        & $A_{\mathrm{Te}_{D_u}} \downarrow$ & $A_{\mathrm{Te}_{D_r}} \uparrow$ & $\mathrm{MIA} \downarrow$ 
        & $A_{{D_u}} \downarrow$ & $A_{\mathrm{Te}_{D_r}} \uparrow$ & $\mathrm{MIA} \downarrow$ \\ 
        \midrule

        \multirow{5}{*}{ResNet18} 
        & Retrain & 34.40±\footnotesize{0.87} & 48.25±\footnotesize{0.09} & 86.40±\footnotesize{0.64} & 0.00±\footnotesize{0.00} & 49.57±\footnotesize{0.18} & 73.17±\footnotesize{0.00}& 33.00±\footnotesize{0.69} & 48.93±\footnotesize{0.10} & 92.67±\footnotesize{0.20} \\ 
        & FT      & 37.20±\footnotesize{1.08} & \textbf{49.61}±\footnotesize{1.05} & 90.53±\footnotesize{0.70} & 3.86±\footnotesize{0.45} & \textbf{50.23}±\footnotesize{0.60} & 85.47±\footnotesize{0.91} & 38.33±\footnotesize{0.99} & \textbf{50.19}±\footnotesize{0.08} & 97.33±\footnotesize{0.33} \\
        & NoT     & \underline{34.33}±\footnotesize{0.94} & \underline{47.19}±\footnotesize{0.20} & 85.13±\footnotesize{0.67} & 6.17±\footnotesize{0.52} & \underline{49.14}±\footnotesize{0.96} & \underline{71.03}±\footnotesize{2.82} & \underline{33.67}±\footnotesize{1.63} & \underline{48.35}±\footnotesize{0.99} & 92.00±\footnotesize{0.48} \\
        & FUSED   & 33.06±\footnotesize{0.42} & 41.87±\footnotesize{0.91} & \underline{65.73}±\footnotesize{0.59} & 4.53±\footnotesize{0.21} & 47.39±\footnotesize{0.93} & 55.67±\footnotesize{0.21} & 32.67±\footnotesize{1.02} & 45.96±\footnotesize{0.94} & \underline{70.33}±\footnotesize{0.61} \\
        & FedOrtho    & \cellcolor{gray!20}\textbf{26.53}±\footnotesize{0.06} & \cellcolor{gray!20}46.89±\footnotesize{0.14} & \cellcolor{gray!20}\textbf{55.06}±\footnotesize{0.56} & \cellcolor{gray!20}\textbf{0.00}±\footnotesize{0.00} & \cellcolor{gray!20}48.91±\footnotesize{0.05} & \cellcolor{gray!20}\textbf{49.20}±\footnotesize{0.24} & \cellcolor{gray!20}\textbf{32.00}±\footnotesize{0.05} & \cellcolor{gray!20}48.15±\footnotesize{0.08} & \cellcolor{gray!20}\textbf{63.00}±\footnotesize{0.63} \\ 
        \bottomrule
    \end{tabular}
    }
    \label{tab:resnet18_unlearning_iid_non_iid}
\end{table*}

\subsection{Centralized Class-Level Unlearning Experiments}

For centralized class-level unlearning, we carried out experiments using three models (Res18, Res34, and Vgg16) on three major datasets (CIFAR-10, CIFAR-100, and Tiny ImageNet), with comparisons made against mainstream methods. In addition to the aforementioned three metrics, training time was added as an evaluation indicator. The results demonstrate that our proposed method not only guarantees effective forgetting and reliable retention performance but also features extremely short unlearning time, achieving excellent comprehensive performance.

\begin{table*}[htbp]
  \caption{Model Performance Comparison Under the Centralized Class Unlearning Scenario (Forgetting a Single Class, Averaged Over 3 Runs with Different Random Seeds, Res18/Res34/Vgg16, CIFAR-10/CIFAR-100/Tiny ImageNet)}
  \centering
  \fontsize{8pt}{9pt}\selectfont
  \setlength{\tabcolsep}{3.5pt}
  \begin{tabular}{cc|cccc|cccc|cccc}
    \toprule
    \multirow{2}{*}{\textbf{Model}} & \multirow{2}{*}{\textbf{Method}} & \multicolumn{4}{c}{\textbf{CIFAR-10}} & \multicolumn{4}{c}{\textbf{CIFAR-100}} & \multicolumn{4}{c}{\textbf{Tiny ImageNet}} \\
    \cmidrule(lr){3-6} \cmidrule(lr){7-10} \cmidrule(lr){11-14}
    & & $A_{Te_{D_u}}{}_{\downarrow}$ & $A_{Te_{D_r}}{}_{\uparrow}$ & $\mathrm{time}_{\downarrow}$ & $\mathrm{MIA}_{\downarrow}$ & $A_{Te_{D_u}}{}_{\downarrow}$ & $A_{Te_{D_r}}{}_{\uparrow}$ & $\mathrm{time}_{\downarrow}$ & $\mathrm{MIA}_{\downarrow}$ & $A_{Te_{D_u}}{}_{\downarrow}$ & $A_{Te_{D_r}}{}_{\uparrow}$ & $\mathrm{time}_{\downarrow}$ & $\mathrm{MIA}_{\downarrow}$ \\
    \midrule
    \multirow{6}{*}{Res18} 
    & Retrain       & 0.00\% & 92.29\% & 4360.80s  & 0.00\% & 0.00\% & 78.17\% & 5557.26s  & 0.00\% & 0.00\% & 64.20\% & 15935.06s & 0.00\% \\
    & SSD           & $\underline{4.37\%}$ & 89.30\% & 21.60s  & 3.82\% & $\underline{1.67\%}$ & 76.29\% & 21.38s  & 2.67\% & $\textbf{0.00\%}$ & \underline{61.56\%}& 136.77s  & 6.67\% \\
    & EMNI          & 9.76\% & $\textbf{91.80\%}$ & 322.02s   & 8.29\% & 5.00\% & 73.89\% & 335.40s   & 6.00\% & $\underline{4.00\%}$ & 59.43\% & 348.42s  & 10.0\% \\
    & Salun         & $\textbf{0.00\%}$ & 87.45\% & $\underline{20.47s}$ & $\textbf{0.60\%}$ & $\textbf{0.00\%}$ & 73.93\% & $\underline{21.55s}$ & $\textbf{0.00\%}$ & $\textbf{0.00\%}$ & 58.27\% & $\underline{45.79s}$ & 6.67\% \\
    & BTF           & 8.61\% & 88.34\% & 23.26s   & 7.74\% & 2.00\% & $\underline{77.00\%}$ & 24.36s   & 4.67\% & 1.33\% & 59.70\%& 47.52s   & $\underline{4.67\%}$ \\
    & FedOrtho      & \cellcolor{gray!20}$\textbf{0.00\%}$ & \cellcolor{gray!20}$\underline{91.34\%}$ & \cellcolor{gray!20}$\textbf{0.1254s}$ & \cellcolor{gray!20}$\underline{1.82\%}$ & \cellcolor{gray!20}$\textbf{0.00\%}$ & \cellcolor{gray!20}$\textbf{77.45\%}$ & \cellcolor{gray!20}$\textbf{0.1135s}$ & \cellcolor{gray!20}$\underline{2.00\%}$ & \cellcolor{gray!20}$\textbf{0.00\%}$ & \cellcolor{gray!20}$\textbf{63.11\%}$ & \cellcolor{gray!20}$\textbf{0.1244s}$ & \cellcolor{gray!20}$\textbf{2.00\%}$ \\
    \midrule
    \multirow{6}{*}{Res34} 
    & Retrain       & 0.00\% & 93.42\% & 5377.80s & 0.00\% & 0.00\% & 79.00\% & 6875.10s & 0.00\% & 0.00\% & 64.73\% & 17051.00s & 0.00\% \\
    & SSD           & $\underline{1.93\%}$ & 90.91\%& 48.60s  & $\textbf{0.20\%}$ & $\textbf{0.00\%}$ & $\underline{77.87\%}$ & 49.00s  & 2.33\% & $\textbf{0.00\%}$ & \underline{61.39\%}& 317.22s  & $\textbf{0.00\%}$ \\
    & EMNI          & 3.27\% & \underline{91.46\%}& 556.26s   & 8.60\% & $\underline{7.33\%}$ & 75.38\% & 562.20s   & 6.67\% & $\underline{2.67\%}$ & 60.04\% & 583.80s  & 6.00\% \\
    & Salun         & $\textbf{0.00\%}$ & 88.17\% & $\underline{27.57s}$ & $\underline{0.00\%}$ & $\textbf{0.00\%}$ & 76.44\% & $\underline{29.20s}$ & $\textbf{0.00\%}$ & $\textbf{0.00\%}$ & 59.79\% & $\underline{62.04s}$ & $\underline{2.67\%}$ \\
    & BTF           & 4.40\% & 86.92\% & 41.29s   & 8.30\% & 1.67\% & 76.10\% & 41.53s   & $\underline{3.67\%}$ & 0.67\% & 59.64\%& 81.60s   & 4.00\% \\
    & FedOrtho      & \cellcolor{gray!20}$\textbf{0.00\%}$ & \cellcolor{gray!20}\textbf{92.05\%}& \cellcolor{gray!20}$\textbf{0.1342s}$ & \cellcolor{gray!20}$\underline{2.60\%}$ & \cellcolor{gray!20}$\textbf{0.00\%}$ & \cellcolor{gray!20}$\textbf{78.52\%}$ & \cellcolor{gray!20}$\textbf{0.1290s}$ & \cellcolor{gray!20}$\underline{2.67\%}$ & \cellcolor{gray!20}$\textbf{0.00\%}$ & \cellcolor{gray!20}$\textbf{63.68\%}$ & \cellcolor{gray!20}$\textbf{0.1350s}$ & \cellcolor{gray!20}$\underline{2.67\%}$ \\
    \midrule
    \multirow{6}{*}{Vgg16} 
    & Retrain       & 0.00\% & 92.61\% & 3976.71s & 0.00\% & 0.00\% & 74.48\%& 4407.80s  & 0.00\% & 0.00\% & 60.89\% & 10033.68s & 0.00\% \\
    & SSD           & $\textbf{0.00\%}$ & 86.19\% & $\underline{13.83s}$ & 4.20\% & $\textbf{0.00\%}$ & 70.43\%& $\underline{13.72s}$ & $\underline{2.67\%}$ & $\textbf{0.00\%}$ & \textbf{59.95\%}& $\underline{61.48s}$ & 4.67\%\\
    & EMNI          & $\underline{2.47\%}$ & $\underline{90.28\%}$ & 289.81s  & 2.50\%& $\underline{3.33\%}$ & 71.57\%& 300.53s   & 8.33\% & $\underline{2.67\%}$ & 56.38\% & 312.63s   & 5.33\% \\
    & Salun         & $\textbf{0.00\%}$ & 87.81\% & 15.77s   & $\textbf{0.00\%}$ & $\textbf{0.00\%}$ & 71.77\%& 16.19s    & $\textbf{0.00\%}$ & $\textbf{0.00\%}$ & 54.59\% & 33.29s    & $\textbf{1.33\%}$ \\
    & BTF           & 9.13\% & 88.83\% & 17.32s   & 4.33\% & 3.67\% & \underline{72.49\%}& 18.07s    & 5.33\% & 0.67\% & 55.78\% & 38.64s    & 6.67\% \\
    & FedOrtho      & \cellcolor{gray!20}$\textbf{0.00\%}$ & \cellcolor{gray!20}$\textbf{91.32\%}$ & \cellcolor{gray!20}$\textbf{0.0780s}$ & \cellcolor{gray!20}$\underline{0.10\%}$& \cellcolor{gray!20}$\textbf{0.00\%}$ & \cellcolor{gray!20}\textbf{73.65\%}& \cellcolor{gray!20}$\textbf{0.0850s}$ & \cellcolor{gray!20}$\underline{2.67\%}$ & \cellcolor{gray!20}$\textbf{0.00\%}$ & \cellcolor{gray!20}\underline{59.24\%}& \cellcolor{gray!20}$\textbf{0.1020s}$ & \cellcolor{gray!20}$\underline{3.33\%}$ \\ 
    \bottomrule
  \end{tabular}
  \label{tab:class_unlearning_performance}
\end{table*}
\subsection{Case Study}
\label{sec:kernel_heatmap}

\begin{figure*}[t]
    \centering
    \includegraphics[width=1.0\textwidth]{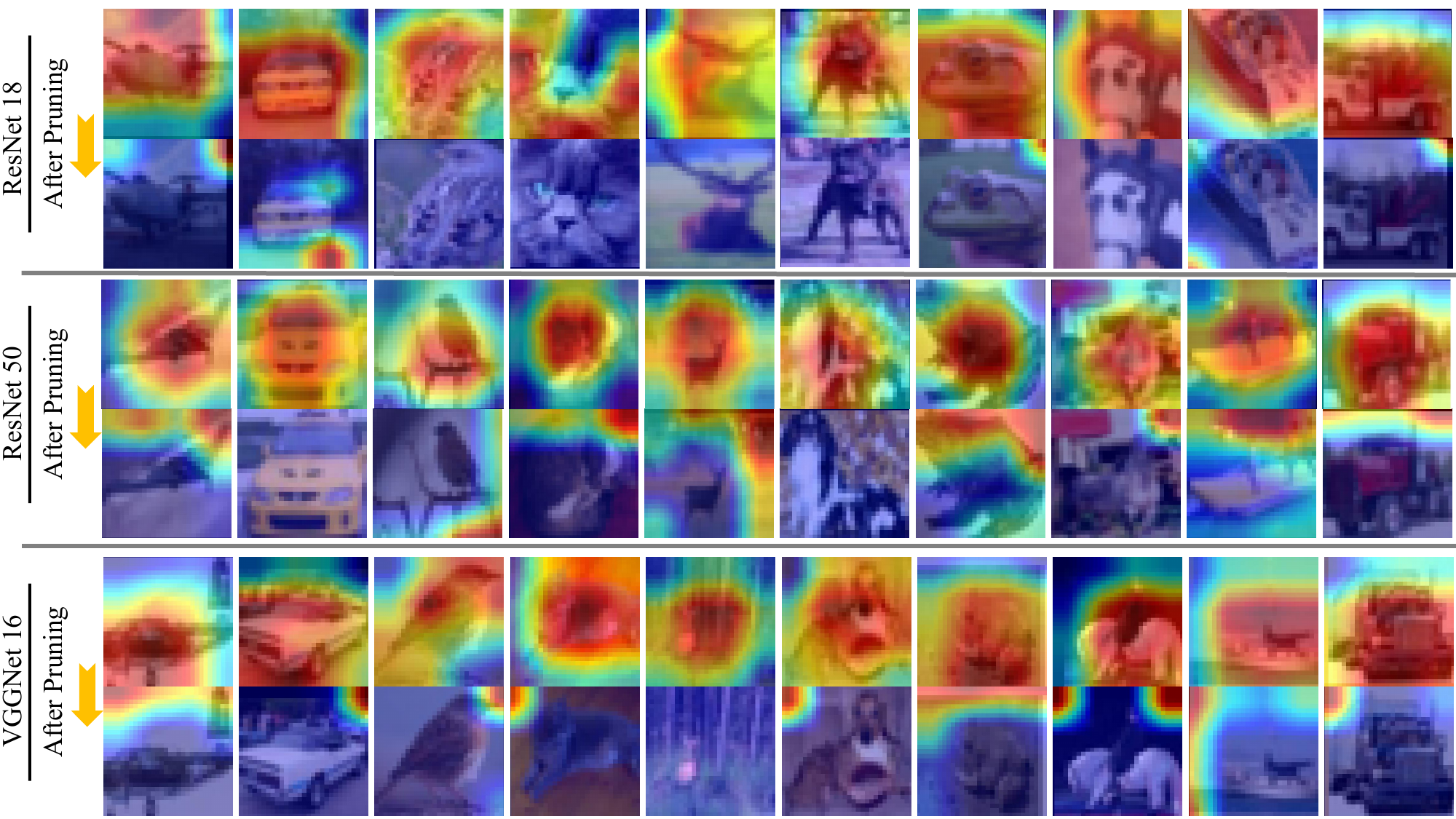}
    \caption{Comparison of convolutional kernel Grad-CAM heatmaps before and after pruning RESNET50, RESNET18, and VGG16 on the CIFAR-10 dataset}
    \label{fig:supply_hotmap}
\end{figure*}

Fig.~\ref{fig:supply_hotmap} presents the Grad-CAM heatmaps of target convolutional kernels for ResNet50, ResNet18, and VGG16 before and after pruning on CIFAR-10. The visualization covers all ten classes, demonstrating that the proposed pruning method can maintain target feature extraction across different network architectures and classes, thus showing excellent generalization ability.

\end{document}